\documentclass[format=acmsmall, review=false, screen=true]{acmart}

\renewcommand\footnotetextcopyrightpermission[1]{} 
\usepackage{subcaption}
\usepackage{tabu}
\usepackage{multirow}

\begin{document}
\title[]{Harvesting Visual Objects from Internet Images via Deep Learning Based Objectness Assessment}

\author{Kan Wu}
\affiliation{
  \department{Department of Computer Science}
  \institution{The University of Hong Kong}
	  \city{Hong Kong}
	  \state{Hong Kong S.A.R., China}
	  }
\email{kwu@cs.hku.hk}

\author{Guanbin Li*}
\affiliation{
  \department{School of Data and Computer Science}
  \institution{Sun Yat-Sen University}
	  \city{Guangzhou}
	  \state{Guangdong, China}
	  }
\email{liguanbin@mail.sysu.edu.cn}

\author{Haofeng Li}
\affiliation{
 \department{Department of Computer Science}
 \institution{The University of Hong Kong}
	 \city{Hong Kong}
	 \state{Hong Kong S.A.R., China}
	 }
\email{lhaof@foxmail.com}

\author{Jianjun Zhang}
\affiliation{
  \department{National Centre for Computer Animation}
  \institution{Bournemouth University}
	  \city{Poole}
	  \state{Dorset, United Kingdom}
	  }
\email{jzhang@bournemouth.ac.uk}

\author{Yizhou Yu}
\affiliation{
  \department{Department of Computer Science}
  \institution{The University of Hong Kong}
 	  \city{Hong Kong}
	  \state{Hong Kong S.A.R., China}
	  }
\affiliation{
  \institution{Deepwise AI Lab}
	  \city{Beijing, China}
	  }
\email{yizhouy@acm.org}

\begin{abstract}
The collection of internet images has been growing in an astonishing speed. It is undoubted that these images contain rich visual information that can be useful in many applications, such as visual media creation and data-driven image synthesis. In this paper, we focus on the methodologies for building a visual object database from a collection of internet images. Such database is built to contain a large number of high-quality visual objects that can help with various data-driven image applications. Our method is based on dense proposal generation and objectness-based re-ranking. A novel deep convolutional neural network is designed for the inference of proposal {\em objectness}, the probability of a proposal containing optimally-located foreground object. In our work, the {\em objectness} is quantitatively measured in regard of {\em completeness} and {\em fullness}, reflecting two complementary features of an optimal proposal: a complete foreground and relatively small background. Our experiments indicate that object proposals re-ranked according to the output of our network generally achieve higher performance than those produced by other state-of-the-art methods. As a concrete example, a database of over 1.2 million visual objects has been built using the proposed method, and has been successfully used in various data-driven image applications.
\end{abstract}


\thanks{*Corresponding author: Guanbin Li}
\thanks{This project was partially supported by the EU H2020 project-AniAge (No.691215) and the National Natural Science Foundation of China (No.61702565).}

\maketitle



\section{Introduction}

Internet images have been growing in an astonishing speed in recent years. With daily increased popularity of online social media sites such as Instagram~\cite{instagram_website}, Flickr~\cite{flickr_website}, Facebook~\cite{facebook_website}, etc., the number of internet images has become quite huge. For example, Flickr, which began in 2005, now hosts more than 13 billion images, and is continuously growing by the minute with more than 120 million active users. The rapid growth of internet images also brings up many applications that make a good use of this big data, such as image retrieval~\cite{image_retrieve-tmm2017-ma}, stitching~\cite{image_stitch-tmm2017-li}, recommendation~\cite{image_recommend-tmm2017-zhang}, image quality assessment~\cite{image_quality_assess-tmm2017-guan}, or benchmark dataset making~\cite{dataset_build_from_web_images-tmm2017-yao}, etc. It is also obvious that within these images the number and richness of contained objects are of great usefulness, especially in some data-driven applications related to object manipulation, such as object retrieval~\cite{obj_retrieval_tmm2011-wang, obj_retrieval2_tmm2011-yang}, classification~\cite{web_object_classification-tmm2013-lu}, enhancement~\cite{image_enhance-tmm2013-zhang}, etc. Traditionally, searching for images that contain desired objects is done through online search engines such as Google Images, which utilizes content-based image retrieval techniques to find images that contain similar objects, or simply acquires images by their tagged information. However, images returned by a search engine can not be easily controlled, as they may be biased towards highly similar objects, or sometimes contain unwanted objects. This observation motivates building a rich object database that is tailored to support various demanding data-driven image applications.

However, mining objects from internet images is challenging. First, information regarding exact object location is not readily available within most internet images. For example, on Flickr~\cite{flickr_website} and Instagram~\cite{instagram_website}, most image tags are ``object types'', ``object names'', ``camera parameters'', `locations where pictures were taken'', etc. Second, there are usually problems like inter-class variation, object occlusion, and background cluttering within images taken by non-professional people~\cite{unsupervise_object_discovery-cvpr2015-cho, regionlets_generic_object_detection-iccv2013-wang}. Third, to collect visual objects, the method should be class-agnostic, without any pre-defined object categories. Currently, there are many methods for locating objects, but most of them assume a predefined, finite set of object categories~\cite{faster_rcnn-nips15-ren}, resulting in limited objects to be discovered. Last but not the least, the object proposals should be of optimal sizes so that foreground objects are properly located within them. In this work, the term {\em objectness} is used to indicate the probability of an object proposal being optimal in location and size. We measure the objectness by two criteria: {\em completeness} and {\em fullness}. {\em Completeness} requires contained foreground to be complete, while {\em fullness} requires the background to be relatively small. Satisfying both {\em completeness} and {\em fullness} is important, especially in some data-driven image applications that require properly-cropped images, such as clip-art illustrations~\cite{photo_clipart-siggraph2007-lalonde}, image compositing~\cite{photo_tourism-siggraph2006-snavely}, imaginary scene synthesis~\cite{sketch2photo-sigraphasia2009-chen}, etc.

In this paper, we propose an object mining pipeline for building a database of objects from internet images. Our pipeline starts with object proposal generation using off-the-shelves methods. Typically for a proposal generator, several partially overlapping proposals are produced around a true object, and they only differ slightly in location, size and aspect ratio. Among these similar proposals, we need to discover the optimal ones that completely cover the objects, since most other proposals either cut off parts of the object (violating {\em completeness}) or contain too much background area (violating {\em fullness}). To this end, we design a deep convolutional neural network for objectness assessment, which is the cornerstone of the whole pipeline. The network architecture is based on the design of two sub-networks for extracting object-oriented features and edge-oriented features separately. For each object proposal box, grid-based spatial pooling is performed on multiple feature maps given by the convolutional layers, resulting in multi-scale aggregated features. Fully-connected layers are used for regression that gives the final objectness score. Quantitative experimental results and user study records indicate that top proposals given by our method exhibit higher performance in general.

In summary, this work has the following contributions:
\begin{itemize}
	\item An effective pipeline for mining objects from internet images. This pipeline is based on dense object proposal generation and objectness assessment.
	\item A novel deep neural network for objectness assessment. The output of this network can be used to choose proposals that have optimal locations and sizes.
	\item A large database of objects that has been successfully used in some data-driven image applications, and can be of potential use in other applications as well, such as image style transferring, object co-segmentation, etc.
\end{itemize}

\section{Related Work}

\subsection{Data-driven Image Editing}

There are many applications that make use of large image collection, such as Lalonde~{\em et al.} 's ``Photo Clip Art''~\cite{photo_clipart-siggraph2007-lalonde}. For a given location where new objects are to be inserted, the system searches within a large database to find objects with desired properties (i.e., category, resolution, camera pose, lighting, etc.). Similar idea is also exploited in scene synthesis, such as Hays~{\em et al.} 's work~\cite{scene_completion-siggraph2007-hays}, where they perform image completion by querying for similar scenes as replacements for selected regions. The rapidly growing collection of internet image also enables synthesizing imaginary scenes using convenient user interaction, such as in ``Sketch2Photo''~\cite{sketch2photo-sigraphasia2009-chen}, where user-drawn sketches are combined with text labels to find suitable background and foreground images for blending. The synthesized images can be quite convincing due to the richness and diversity of internet images that easily guarantee a reasonable combination of found objects and backgrounds. Wang~{\em et al.} 's work~\cite{bigger_picture-siggraphasia2014-wang} uses internet images on image extrapolation, where candidate contents near edges are obtained from a pre-built database using graph-matching over hierarchically-segmented patches. Aside from that, large image collection can also be useful in style transferring and colorization, such as Tsai~{\em et al.} 's automatic sky replacement algorithm~\cite{Tsai_SIGGRAPH_2016} that generates naturally looking images with various sky styles, and Chia~{\em et al.} 's image colorization system~\cite{chia2011semantic} that leverages the rich image contents from the internet. On the other hand, for exploring and visualizing a large set of images of the same scene, Snavely {\em et al.} present ``Photo Tourism'', a system that recovers 3D viewing scene geometry by matching dense key feature points among photos taken from different angles~\cite{photo_tourism-siggraph2006-snavely}. Russell {\em et al.} propose to organize online text information to explore images within a reconstructed virtual scene~\cite{3d_wikipedia-siggraphasia2013-russell}.

A common key component within many data-driven image editing systems is a large image database that contains significantly diversified visual contents. More importantly, the images within should also be of good quality to be useful in real applications. Regarding object-oriented image editing applications, such as ``Photo Clip Art''~\cite{photo_clipart-siggraph2007-lalonde} and ``Sketch2Photo''~\cite{sketch2photo-sigraphasia2009-chen}, the object images within the database should have two important features: complete and clear foregrounds and less interfering backgrounds. In multimedia area, there are many works related to material quality assessment, such as visual importance and distortion evaluation on images~\cite{image_quality_assess-tmm2017-guan} and quality-of-experience rating on videos~\cite{deep_qoe-icme2018-zhang}. In this work, the ``quality'' to be assessed is different. The goal is to quantitatively measure image objectness that indicates the presence of foreground objects, instead of color, distortion, resolution, frame rate, etc.

\subsection{Object Proposals and Objectness}
For a given image, object proposals are referred to as bounding boxes that are likely to contain meaningful objects. The term {\em objectness} is usually used to indicate the probability of the presence of a foreground object~\cite{deep_contrast-tnnls2018-li}. Object proposals are usually generated by window sampling, followed by some searching or rating process to further narrow down optimal boxes to keep~\cite{selective_search-ijcv2013-uijlings}. It is straight-forward to use exhaustive searching over entire sampling space to obtain proposal windows~\cite{exhaustive_search_proposal-iccv2009-harzallah}. However, exhaustive searching is usually computationally expensive, forcing coarse sampling grids or low-level image features to be used in objectness calculation~\cite{selective_search-ijcv2013-uijlings}. Therefore, most state-of-the-art methods exploit more adaptive methods for searching proposal windows, such as randomly picking proposals with high object classification scores~\cite{alexe2012measuring}, or merging image segments obtained from hierarchical segmentation using certain strategies~\cite{selective_search-ijcv2013-uijlings, mcg-cvpr2014-arbelaez, cob-eccv2016-maninis, adaptive_distance_metric-cvpr2015-xiao},~etc. The measurement of objectness has been widely explored in object proposal generators that aim at locating target objects through a minimum number of proposal window hypotheses~\cite{alexe2012measuring}. In Alexe~{\em et al.} 's work~\cite{alexe2012measuring}, multiple measurements of objectness such as multi-scale saliency, edge density, super pixels straddling, ~etc., are combined in a Bayesian framework to obtain final objectness scores. In Selective Search~\cite{selective_search-ijcv2013-uijlings}, hierarchical segmentation results are merged in a bottom-up fashion by similarities between adjacent regions. In MCG~\cite{mcg-cvpr2014-arbelaez}, proposal windows are merged hierarchically using combinatorial grouping that maintains high achievable performance and are further sorted by a regression model trained on hierarchical features that can be computed efficiently. In Edge Box~\cite{edgebox-eccv2014-zitnick}, the number of contours that are completely contained in a bounding box is used as an indication of objectness. Other than directly rating proposals, Lu {\em et al.} 's work uses a different strategy of rejecting bounding boxes that have no explicit closed contours~\cite{contour_box-iccv2015-lu}.

Recently, deep convolutional neural networks~(CNNs) have achieved great successe in image recognition~\cite{krizhevsky2012imagenet, vgg16-corr2014-simonyan} and object detection~\cite{faster_rcnn-nips15-ren, redmon2015you}. There are also successful attempts to apply deep CNNs to object proposal generation, such as Kuo {\em et al.} 's Deep Box~\cite{deep_box-iccv2015-kuo} and Ghodrati {\em et al.} 's Deep Proposal~\cite{deep_proposal-iccv2015-ghodrati}. Recently, Ren {\em et al.} introduce a region proposal network to predict object bounding boxes and objectness scores simultaneously~\cite{faster_rcnn-nips15-ren}. Researchers have also been using neural networks in saliency detection~\cite{mdf-tip2016-li, deep_contrast-tnnls2018-li}, which can also be used in measuring objectness in object proposal generation and instance-level segmentation~\cite{instance_level_saliency_object_segmentation-cvpr2017-li}. However, foreground objects in images are not always ``salient'' in images. There are also other works that apply fully convolutional networks to infer class-agnostic segmentation masks, along with their objectness scores~\cite{deep_mask-cvpr2015-pedro, sharp_mask-cvpr2016-pedro,dai2016instance}. More detailed discussion and evaluation on object proposal generators can be found in~\cite{Chavali_2016_CVPR}. There are also some works that exploit the richness of internet images for locating objects, such as Tang {\em et al.} 's object co-localization~\cite{Tang_2014_CVPR}, or the detection models trained with internet images, such as those in Chen {\em et al.} 's work~\cite{NEIL-ICCV2013-Chen} and Divvala 's work~\cite{Divvala_2014_CVPR}.

In general, many object proposal generators are designed to serve as pre-processing tools for object detection which aims at locating objects by specific category labels. Achieving a high object recall is thus more important than the accuracy of proposal windows, as missing objects in this stage cannot be rediscovered in the next stage of object detection. However, from most recent successful data-driven image applications, we see an increasing requirement on the quality of the database, that accuracy of the detected objects is relatively more important than recall. Using most existing methods, proposals that are chosen may enclose incomplete objects or too much background contents that can interfere with critical image editing operations like segmentation, blending, etc.


\section{Overview}\label{sec:overview}

\begin{figure*}
	\centering
	\includegraphics[width=1.0\linewidth]{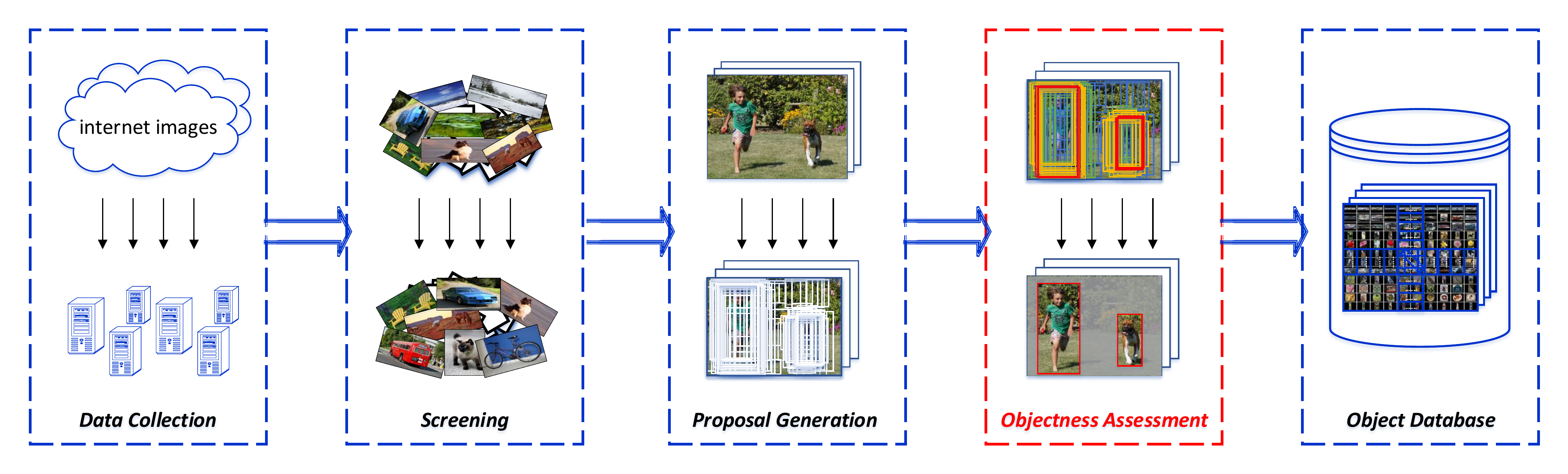}
	\caption{The overall object mining pipeline.}
  \label{fig:01_pipeline}
\end{figure*}

Our pipeline for building an object database is illustrated in Figure~\ref{fig:01_pipeline}. We first collect images from a widely-used photo sharing website, Flickr~\cite{flickr_website}. All downloaded images go through a screening process that discards images with no clean foreground objects. The screening is done by applying a binary classifier that separates desired images from the rest. The next step is the generation of object proposals from all remaining images. We leave this task to many state-of-the-art methods such as MCG~\cite{mcg-cvpr2014-arbelaez}, RPN~\cite{faster_rcnn-nips15-ren}, or COB~\cite{cob-eccv2016-maninis}, etc. The proposal objectness assessment step that comes after (enclosed by red rectangle in Figure~\ref{fig:01_pipeline}) is one of the major contributions of this paper. All object proposals are rated with objectness scores by a deep convolutional neural network that makes inferences on multi-scale object-oriented and edge-oriented features. After objectness assessment, we screen the object proposals by thresholding with respect to their scores, then sort the remaining proposals in descending order. A non-maximal suppression (NMS) is then used to reduce duplicated proposal boxes. Object proposals surviving all the above steps are used to build the database, which now contains more than 1.2 million individual visual objects.

The paper is organized as followed. Section~\ref{sec:collection} describes image collection and pre-processing steps. Section~\ref{sec:prop-assess} defines network architecture and training process. Section~\ref{sec:evaluation} reports experimental results related to performance of the network. Section~\ref{sec:app} lists some data-driven applications that can make use of the database.


\section{Image Collection and Pre-processing}\label{sec:collection}

In this work, images are collected from Flickr. We search for images using preset 36 city geolocations, approximate search radius of 20 km and timespan between January 2005 and January 2015. To respect the rights and privacy of other internet users, we only download images licensed with the permission to redistribute derived works, such as the license of {\em Creative Commons}. The result of this process is a collection of approximately 3 million images, taking storage space of 300GB. These images should cover good geographical and cultural diversities around the world.

However, not all downloaded images can be useful for image editing applications. Some of the images, such as photos of the sky and distant mountains, appear to contain no meaningful objects. Images with crowded people, where severe occlusion often occur, can hardly be useful either. To reduce unnecessary processing that will not produce useful objects, we perform a screening process using a binary classification network that decides whether to keep a given image. The structure of the network is the same as the VGG-16 network for image classification~\cite{vgg16-corr2014-simonyan} except that the number of neurons in the original softmax layer is changed to two, giving two probabilistic outputs ({\em positive} and {\em negative}) that indicate whether or not to keep an image. The weights of original VGG-16 network pre-trained on ImageNet are used to initialize this network, and further fine-tuned with our own training data. The training data consists of approximately 5000 images, hand-picked from the training sets of ImageNet~\cite{krizhevsky2012imagenet}, PASCAL VOC2007~\cite{voc_2007} and PASCAL VOC2012~\cite{voc_2012}. We label half of these images as positive (having clear foreground objects) and the other half as negative (not having clear foreground objects). We remove those images predicted as negative by this binary classifier. After this screening step, we have approximately 1/3 of originally downloaded images left. It is possible that the binary classification can be over-fitting. However, in this work, the accuracy of the binary classifier is not important. The key point of this step is to remove as many useless images as possible, while the objects in some falsely rejected images can be easily compensated by those in other images since the whole collection of images is huge.

The space of all possible window locations within an image can be huge. Instead of directly searching for optimal windows, we use off-the-shelves object proposal generators for proposal box generation. Several methods are tried, including Selective Search~\cite{selective_search-ijcv2013-uijlings}, Edge Box~\cite{edgebox-eccv2014-zitnick}, MCG~\cite{mcg-cvpr2014-arbelaez}, RPN from Faster R-CNN~\cite{faster_rcnn-nips15-ren}, and Deep Proposal~\cite{deep_proposal-iccv2015-ghodrati}. In this work, we prefer fast proposal generation methods with good precision so that later our assessment network can indeed find optimal object proposals. It is worth mentioning that in this step, precision is relatively more important than recall since our method only re-ranks existing proposals. The problem of relatively low recall can be ignored due to the huge number of internet images, while poor precision will make the entire pipeline less effective. In our experiments, we use MCG~\cite{mcg-cvpr2014-arbelaez}, COB~\cite{cob-eccv2016-maninis} and RPN~\cite{faster_rcnn-nips15-ren} due to their relatively good precision and execution efficiency.

\section{Objectness Assessment Network}\label{sec:prop-assess}

We design a regression network that rates proposals based on objectness. For any given image and its proposal windows, the network will output scores that indicate probabilities of proposals being optimal in regard of enclosed objects. Note that the whole pipeline assumes that there are optimal proposals being produced by proposal generators, as our network does not generate new object proposals. This is actually a reasonable assumption due to the existence of many good object proposal generators~\cite{selective_search-ijcv2013-uijlings, mcg-cvpr2014-arbelaez, deep_box-iccv2015-kuo,cob-eccv2016-maninis, faster_rcnn-nips15-ren}. The only problem is to find optimal ones among all generated proposals.

\subsection{Objectness Assessment Criteria}\label{subsec:quality_assessment_criteria}

The optimality of a given object proposal window is quantitatively measured with respect to two previously mentioned criteria: {\em completeness} and {\em fullness}. {\em Completeness} requires a foreground object to be entirely located within the proposal window, while {\em fullness} requires the background area to be relatively small compared to the foreground.

Denote proposal window as $P$, corresponding ground-truth box as $G$, and their intersected region as $I$. The {\em completeness index} $C_c$ and {\em fullness index} $C_f$ are calculated as Equation~\ref{eq:completeness_fullness_index},

\begin{equation}\label{eq:completeness_fullness_index}
	\begin{matrix}
		C_c=Area(I)/Area(G)
		\\
		C_f=Area(I)/Area(P)
	\end{matrix}
\end{equation}

\begin{figure}
	\centering
	\includegraphics[width=0.55\linewidth]{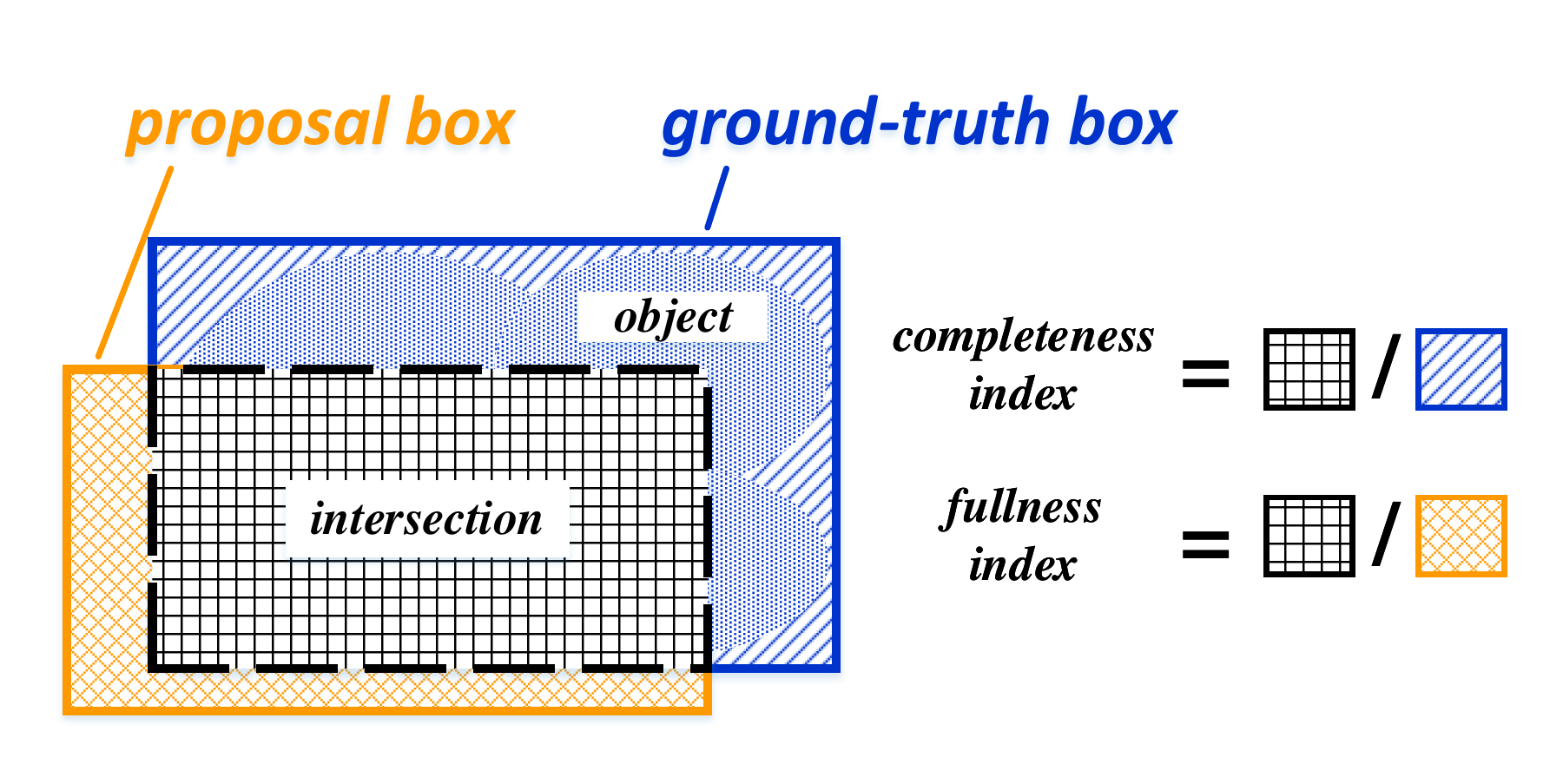}
	\caption{Calculation of completeness index and fullness index.}
	\label{fig:calc_completeness_fullness_index}
\end{figure}

\begin{figure*}
	\centering
	
	\begin{subfigure}[]{0.45\linewidth}
		\includegraphics[width=\linewidth]{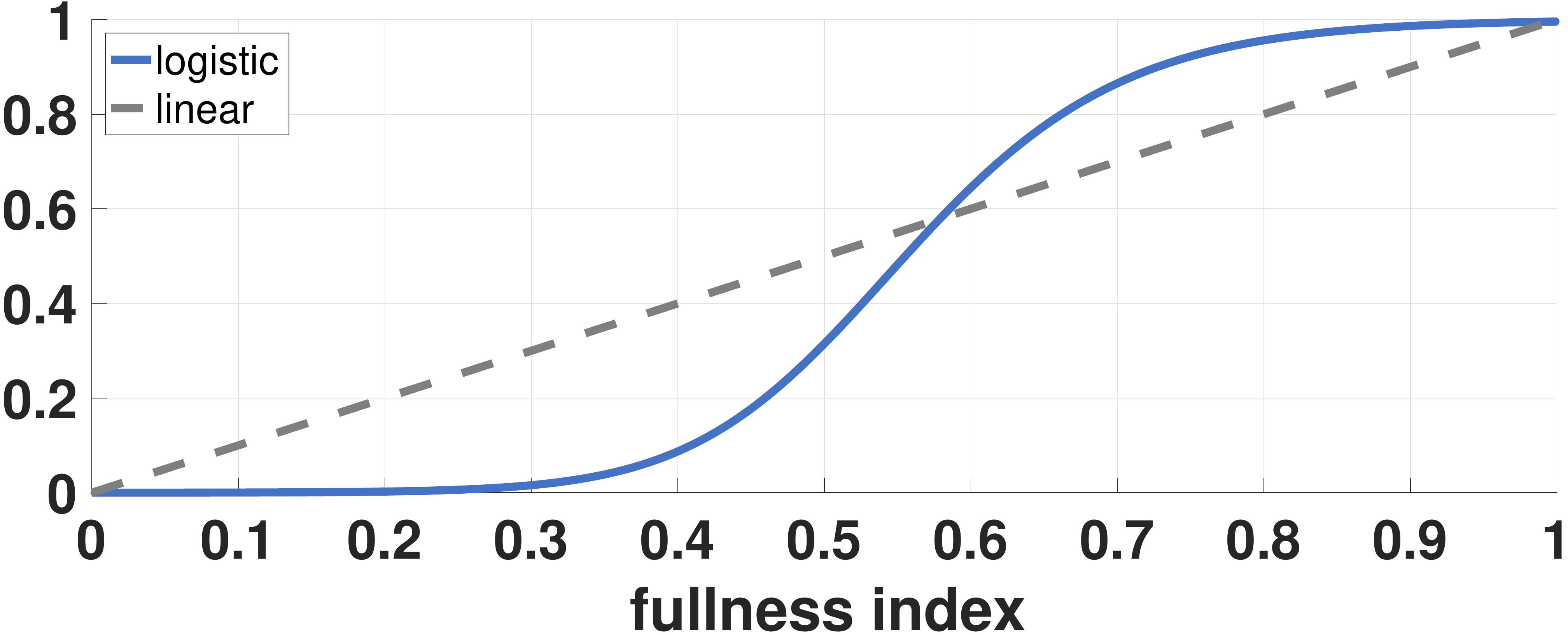}
		\caption{Calculation of fullness score}
		\label{fig:fullness_score}
	\end{subfigure}
	\begin{subfigure}[]{0.45\linewidth}
		\includegraphics[width=\linewidth]{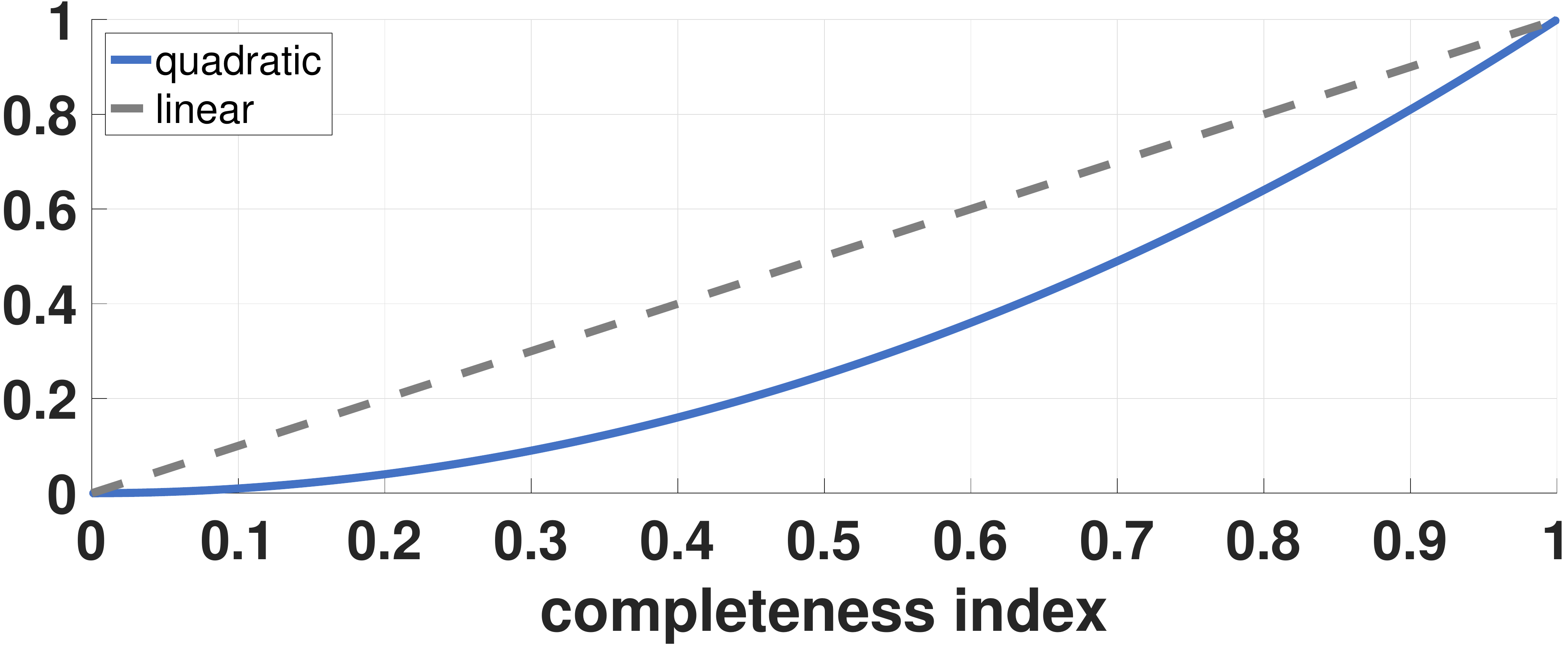}
		\caption{Calculation of completeness score}
		\label{fig:completeness_score}
	\end{subfigure}

	\caption{Calculation of completeness score and fullness score. Completeness index and fullness index are calculated according to methods illustrated in Figure~\ref{fig:calc_completeness_fullness_index}. Fullness score functions correspond to Equation~\ref{eq:fullness_score_logistic}, where $\alpha=0.5$, $\beta=12$, $\gamma=0.6$, $q=1$. Completeness score function corresponds to Equation~\ref{eq:completeness_score_quadratic}.}
	\label{fig:calc_completeness_fullness_score}
\end{figure*}

where $Area(P)$, $Area(G)$, $Area(I)$ are the area of proposal window, ground-truth box and intersected box, respectively. The calculation is also illustrated in Figure~\ref{fig:calc_completeness_fullness_index}. For both $C_c$ and $C_f$, the more intersected area within the ground-truth box, the higher quantity they take, with the worst case being both terms take $0$ for no intersection, and the best case that both terms are $1$ for complete overlapping. However, although their calculation is similar, these two terms measure different aspects of proposal optimality and complement each other in various cases. For example, when the proposal box entirely encloses the ground-truth box, $C_c=1$, indicating good completeness, while $C_f$ measures the percentage of area within proposal box that is occupied by an actual object. Similarly, when the ground-truth box entirely encloses the proposal box, $C_f=1$, meaning good fullness, while $C_c$ measures how much of the actual object is enclosed by the proposal window. The separated calculation of {\em completeness index} and {\em fullness index} also enables handling {\em completeness} and {\em fullness} differently by using different ``transfer functions'' on them.

The {\em completeness score} and {\em fullness score} are computed based on previously obtained {\em completeness index} and {\em fullness index}, by using ``transfer functions'' to control the contribution of each index to the score. The calculation is illustrated in Figure~\ref{fig:calc_completeness_fullness_score}. {\em Completeness score} and {\em fullness score} are denoted as $S_c$ and $S_f$, respectively. For {\em fullness score}, we are relatively tolerant when there is only a small background area within proposal window, but make the score drop quickly when the index is below certain threshold. Generalized logistic functions are ideal for this purpose. We use a simplified logistic function,

\begin{equation}\label{eq:fullness_score_logistic}
S_{f}=\frac{1}{{1+qe^{-(\beta(C_f-\alpha)))}}^{1/\gamma}}
\end{equation}

where $\alpha$ and $\beta$ are scaling parameters that map the desired portion of logistic function to the range of $0.0-1.0$. $\gamma$ controls the slope of the curve, and $q$ controls the threshold where the curve starts to drop rapidly. Figure~\ref{fig:fullness_score} illustrates the logistic function we use. It can be seen in Figure~\ref{fig:fullness_score} that $q$ generally controls the tolerant threshold of {\em fullness index}. The score drops relatively slower when $C_f$ is near $1$, but decreases quickly to $0$ once $C_f$ is below the approximate threshold of $0.5$. In all our experiments, $\alpha=0.5$, $\beta=12$, $\gamma=0.6$, $q=1$.

As for {\em completeness score}, we prefer to be less tolerant about objects being cut off by proposal window borders. Thus, we let {\em completeness score} drop rapidly even when a small part of the annotated object is cut off. This can be done by quadratic function,

\begin{equation}\label{eq:completeness_score_quadratic}
S_c={C_c}^2
\end{equation}

It can be seen in Figure~\ref{fig:completeness_score} that near $1.0$ even a small dropping of {\em completeness index} will result in large decreasing of {\em completeness score}. This enforces a large penalty for missing object parts within the proposal window.

The final ground-truth objectness score is a weighted score of {\em completeness score} and {\em fullness score}:

\begin{equation}\label{eq:final_quality_score}
S_{final}=wS_c+(1-w)S_f
\end{equation}

where $w$ is a weight parameter. It generally reflects how we treat a proposal box regarding the satisfaction of ``completeness'' and ``fullness''. Smaller values of w indicate that we value ``fullness'' more over ``completeness'', and vice versa. $w$ is tuned by a training process described in Section~\ref{subsec:scheme_optimize}.

\begin{figure*}
	\centering
	\includegraphics[width=0.9\linewidth]{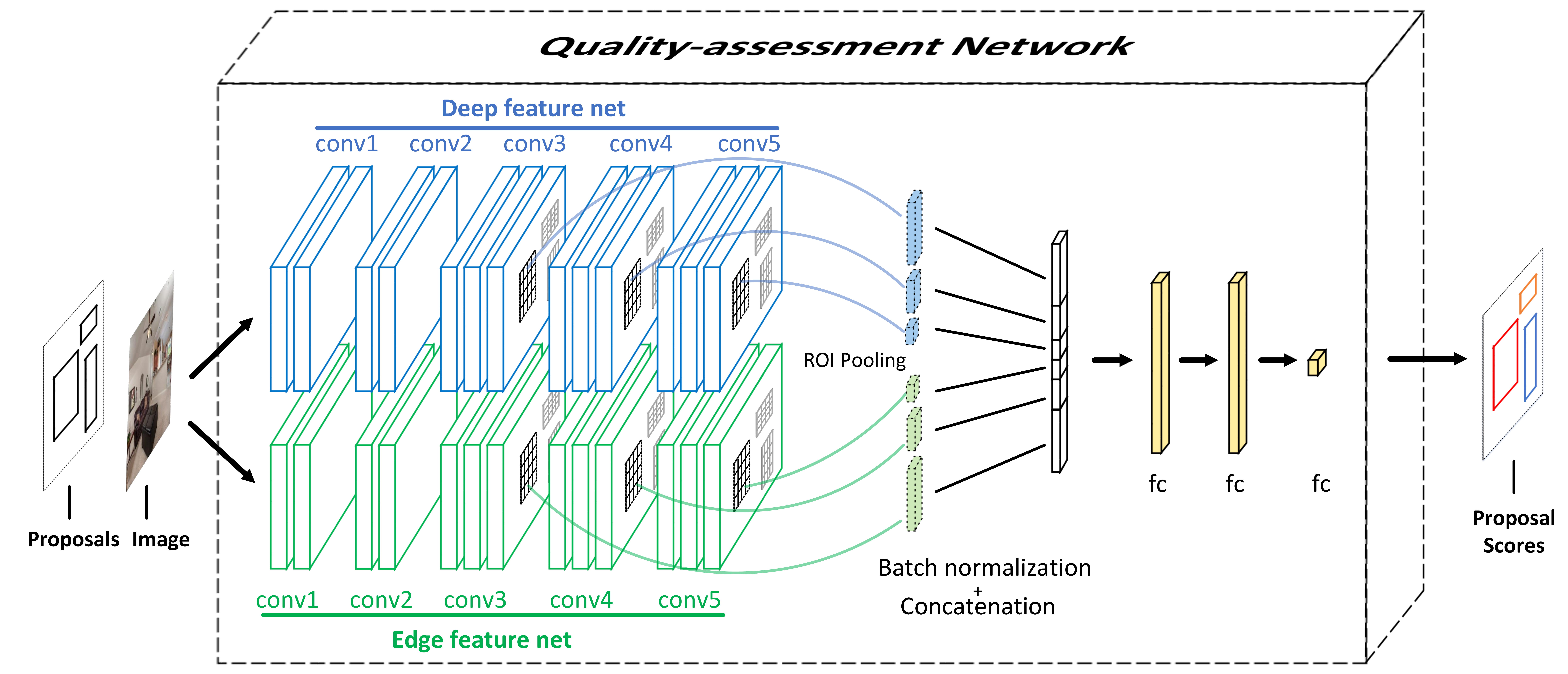}
	\caption{Architecture of our proposal objectness assessment network. Among the convolutional layers, blue ones and green ones are respectively for object-oriented and edge-oriented feature extraction. The input are an image and its associated object proposals, and the output are scalar value scores corresponding to proposal objectness.}
	\label{fig:network_archiecture}
\end{figure*}

\begin{table*}
	\centering
	\huge
	\caption{The configuration of our network. Note that both sub-networks (green and blue parts in Figure~\ref{fig:network_archiecture}) have the same structure of convolutional and pooling layers (the groups of ``conv1'', ``conv2'', ``conv3'', ``conv4'', ``conv5'' below)}
	\label{tab:network_config}
	\resizebox{1.0\textwidth}{!}
	{
		\begin{tabu}{|[2pt]c|[2pt]l|l|l|l|[2pt]l|l|l|l|[2pt]}

		\tabucline[2pt]{-}
		
		\multirow{2}{*}{\textbf{GROUP}} & \multirow{2}{*}{\textbf{LAYER}} & \multirow{2}{*}{\textbf{TYPE}} & \multirow{2}{*}{\textbf{OUTPUT}} & \multirow{2}{*}{\textbf{PARAMETER}} & \multirow{2}{*}{\textbf{LAYER}} & \multirow{2}{*}{\textbf{TYPE}} & \multirow{2}{*}{\textbf{OUTPUT}} & \multirow{2}{*}{\textbf{PARAMETER}}\\
		{} & {} & {} & {} & {} & {} & {} & {} & {}\\
		
		\tabucline[2pt]{-}
		
		input & image & Input & $224\times224\times3\times1$ & - & boxes & Input & $4\times256\times1\times1$ & -\\
		
		\tabucline[2pt]{-}
		
		\multirow{3}{*}{conv1$(\times2)$} &
			conv1\_1 & Convolution & $224\times224\times64\times1$ & kernel\_size=3, filter\_num=64, pad=1 & relu1\_1 & ReLU & $224\times224\times64\times1$ & -\\
		\tabucline[1pt]{2-9}
		{} & conv1\_2 & Convolution & $224\times224\times64\times1$ & kernel\_size=3, filter\_num=64, pad=1 & relu1\_2 & ReLU & $224\times224\times64\times1$ & -\\
		\tabucline[1pt]{2-9}
		{} & pool1 & Pooling & $112\times112\times64\times1$ & type=max, kernel\_size=2, stride=2 & - & - & - & -\\
		
		\tabucline[2pt]{-}
		
		\multirow{3}{*}{conv2$(\times2)$} & 
			conv2\_1 & Convolution & $112\times112\times128\times1$ & kernel\_size=3, filter\_num=128, pad=1 & relu2\_1 & ReLU & $112\times112\times128\times1$ & -\\
		\tabucline[1pt]{2-9}
		{} & conv2\_2 & Convolution & $112\times112\times128\times1$ & kernel\_size=3, filter\_num=128, pad=1 & relu2\_2 & ReLU & $112\times112\times128\times1$ & -\\
		\tabucline[1pt]{2-9}
		{} & pool2 & Pooling & $56\times56\times128\times1$ & type=max, kernel\_size=2, stride=2 & - & - & - & -\\
		
		\tabucline[2pt]{-}
		
		\multirow{5}{*}{conv3$(\times2)$} &
			conv3\_1 & Convolution & $56\times56\times256\times1$ & kernel\_size=3, filter\_num=256, pad=1 & relu3\_1 & ReLU & $56\times56\times256\times1$ & -\\
		\tabucline[1pt]{2-9}
		{} & conv3\_2 & Convolution & $56\times56\times256\times1$ & kernel\_size=3, filter\_num=256, pad=1 & relu3\_2 & ReLU & $56\times56\times256\times1$ & -\\
		\tabucline[1pt]{2-9}
		{} & conv3\_3 & Convolution & $56\times56\times256\times1$ & kernel\_size=3, filter\_num=256, pad=1 & relu3\_3 & ReLU & $56\times56\times256\times1$ & -\\
		\tabucline[1pt]{2-9}
		{} & pool3 & Pooling & $28\times28\times256\times1$ & type=max, kernel\_size=2, stride=2 & - & - & - & -\\
		\tabucline[1pt]{2-9}
		{} & roi\_pool3 & ROI Pooling & $7\times7\times256\times256$ & type=max, grid\_size=7 & - & - & - & -\\
		
		\tabucline[2pt]{-}
		
		\multirow{5}{*}{conv4$(\times2)$} &
			conv4\_1 & Convolution & $28\times28\times512\times1$ & kernel\_size=3, filter\_num=512, pad=1 & relu4\_1 & ReLU & $28\times28\times512\times1$ & -\\
		\tabucline[1pt]{2-9}
		{} & conv4\_2 & Convolution & $28\times28\times512\times1$ & kernel\_size=3, filter\_num=512, pad=1 & relu4\_2 & ReLU & $28\times28\times512\times1$ & -\\
		\tabucline[1pt]{2-9}
		{} & conv4\_3 & Convolution & $28\times28\times512\times1$ & kernel\_size=3, filter\_num=512, pad=1 & relu4\_3 & ReLU & $28\times28\times512\times1$ & -\\
		\tabucline[1pt]{2-9}
		{} & pool4 & Pooling & $14\times14\times512\times1$ & type=max, kernel\_size=2, stride=2 & - & - & - & -\\
		\tabucline[1pt]{2-9}
		{} & roi\_pool4 & ROI Pooling & $7\times7\times512\times256$ & type=max, grid\_size=7 & - & - & - & -\\
		
		\tabucline[2pt]{-}
		
		\multirow{5}{*}{conv5$(\times2)$} &
			conv5\_1 & Convolution & $14\times14\times512\times1$ & kernel\_size=3, filter\_num=512, pad=1 & relu5\_1 & ReLU & $14\times14\times512\times1$ & -\\
		\tabucline[1pt]{2-9}
		{} & conv5\_2 & Convolution & $14\times14\times512\times1$ & kernel\_size=3, filter\_num=512, pad=1 & relu5\_2 & ReLU & $14\times14\times512\times1$ & -\\
		\tabucline[1pt]{2-9}
		{} & conv5\_3 & Convolution & $14\times14\times512\times1$ & kernel\_size=3, filter\_num=512, pad=1 & relu5\_3 & ReLU & $14\times14\times512\times1$ & -\\
		\tabucline[1pt]{2-9}
		{} & roi\_pool5 & ROI Pooling & $7\times7\times512\times256$ & type=max, grid\_size=7 & - & - & - & -\\

		\tabucline[2pt]{-}
		
		\multirow{5}{*}{fc} & concat & Concatenation & $7\times7\times1280\times256$ & axis=1 & - & - & - & -\\
		\tabucline[1pt]{2-9}
		{} & fc1 & Fully-connected & $400\times256\times1\times1$ & neuron\_num=400 & bn\_1 & Batch-normalization & $400\times256\times1\times1$ & -\\
		\tabucline[1pt]{2-9}
		{} & nn\_relu1 & ReLU & $400\times256\times1\times1$ & - & dropout1 & Dropout & $400\times256\times1\times1$ & -\\
		\tabucline[1pt]{2-9}
		{} & fc2 & Fully-connected & $400\times256\times1\times1$ & neuron\_num=400 & bn\_2 & Batch-normalization & $400\times256\times1\times1$ & -\\
		\tabucline[1pt]{2-9}
		{} & nn\_relu2 & ReLU & $400\times256\times1\times1$ & - & dropout2 & Dropout & $400\times256\times1\times1$ & -\\
		\tabucline[2pt]{-}
		
		\multirow{2}{*}{output} &
			fc3 & Fully-connected & $1\times256\times1\times1$ & neuron\_num=1 & bn\_2 & Batch-normalization & $1\times256\times1\times1$ & -\\
		\tabucline[1pt]{2-9}
		{} & nn\_relu3 & ReLU & $1\times256\times1\times1$ & - & - & - & - & -\\
		
		\tabucline[2pt]{-}

		\end{tabu}
	}
\end{table*}

\subsection{Network Architecture}\label{sec:network_architecture}

We aim to design an end-to-end assessment method as a deep regression network, which takes an image and its associated object proposals as input, and predicts objectness scores on the scale of $0.0-1.0$. These scores should reflect how likely each of the proposal windows is in optimal location. The output of the network is a vector, with each element corresponding to one score of an input object proposal.

To conceive such an end-to-end architecture, we have the following considerations. First, the network should utilize both object-oriented and edge-oriented features, as we believe both region and edge features to be important clues for optimal proposal box locations. Object-oriented features can tell whether a proposal window contains an object while edge-oriented features provide additional evidence related to completeness and fullness (for example, a salient edge intersecting the border of a proposal window is a clear indication of missing some object parts). Second, the network should consider high-level features as well as mid-level features when inferring objectness as the former supplies semantic information at the scale of whole objects while the latter supplies semantic information at the scale of object parts. Last but not the least, the network should be very efficient and end-to-end trainable, meaning that only necessary computation is performed, and the interface for applications should be simple.

As shown in Figure~\ref{fig:network_archiecture}, our proposed network architecture is composed of two parallel convolutional layer groups for feature extraction, and two fully connected layers for objectness score regression. We modify the original VGG network by removing its fully connected layers and attaching an ROI pooling layer to each of the last three convolutional layers. An ROI pooling layer basically performs single-scale spatial pyramid pooling~\cite{sppnet_cvpr2015_he} that divides each region of interest into a uniform grid of $H\times W$ cells and performs pooling independently within each grid cell. In this work, we use the grid size of $7\times 7$ for ROI pooling, following Faster R-CNN~\cite{faster_rcnn-nips15-ren}. The outcome of ROI pooling is a fixed-length feature vector, which is the concatenation of max-pooling responses at all grid cells.

The two modified VGG networks in our architecture are used for extracting different types of features. One is  for object-oriented features (blue parts in Figure~\ref{fig:network_archiecture}), the other is for edge-oriented features (green parts in Figure~\ref{fig:network_archiecture}). It is important to include edge detection layers (green layers in Figure~\ref{fig:network_archiecture}) as features extracted with the aim of edge detection contain information of potential object boundary locations, which is closely related to the criteria of {\em completeness} and {\em fullness} and the inference of the final objectness score. For each proposal window, we concatenate the results of six ROI pooling layers and feed them to two fully connected layers for final score prediction. Note that one forward pass of our network can compute the scores for all proposal windows associated with a single input image. This architecture maximizes the throughput of the network, avoids redundant feature map computation, and provides simple interface for practical applications.

Note that a modified VGG with an ROI pooling layer has been used in \cite{faster_rcnn-nips15-ren}. However, in their work ROI pooling is only applied to a single convolutional layer (``conv5\_3''), followed by a simple softmax layer. Although the features computed from ``conv5\_3'' can well characterize semantic information, it is of very low resolution and cannot identify proposal bounding boxes with high-precision. Table~\ref{tab:network_config} gives the configuration of all layers in our network.

\subsection{Network Training}\label{sec:network_training}

The training data for our proposal assessment network includes a set of training images along with a list of annotated ground-truth bounding boxes associated with training images. Each ground-truth bounding box optimally encloses a single visual object. For each training image, we empirically generate 1000 object proposals using an object proposal generator. The ground-truth scores of these object proposals are calculated according to methods described in Section~\ref{subsec:quality_assessment_criteria}. In practice, for each object proposal, we compute its score with respect to every annotated ground-truth bounding box within the image and take the highest one as its final ground-truth score.

The loss function during network training is defined as the mean squared difference between predicted scores and ground-truth scores, averaged over a training batch. Given a mini-batch containing $N$ proposal windows, the training loss is computed as,

\begin{equation}\label{eq:training_loss}
	L=\frac{1}{2N}\sum_{i=1}^{N}\left\|S_{i}^{p}-S_{i}^{g}\right\|_{2}^{2}
\end{equation}

where $S_{i}^{p}$ and $S_{i}^{g}$ are respectively the predicted score and ground-truth score of the $i$-th proposal window within the batch. $S_{i}^{g}$ is calculated according to Equation~\ref{eq:final_quality_score}.

In this work, all experiments are carried out using Caffe~\cite{caffe-cvpr2014-jia}, with a few of our own customized layers. Note that the assessment of object proposals has stringent criteria about {\em completeness} and {\em fullness}, and is generally more challenging than conventional object detection, which only requires a positive proposal window to have more than 50\% overlap with a ground-truth bounding box. Yet, we only have limited training data.

Instead of training the entire network from scratch, we make use of existing models by fine-tuning their pre-trained weights. Two pre-trained models are used in this work. One is for object detection~\cite{faster_rcnn-nips15-ren}, and the other is for edge detection~\cite{hed-iccv2015-xie}. We only take the weights of the convolutional layers in these two pre-trained networks, and the weights of fully connected regression layers are randomly initialized. In the beginning, all weights in the convolutional layers and fully connected layers are trained with respect to the training loss of Equation~\ref{eq:training_loss}. However, through our experiments, we find out that fine-tuning convolutional layers along with fully connected layers does not give better performance. In fact, as the result of our experiments suggests, tuning convolutional layers results in slightly worse performance. On VOC2012, approximately 0.6\% worse precision and 0.3\% worse recall are observed. We believe the small drop of performance is due to some over-fitting in convolutional layers that worsen the feature extraction capabilities on different groups of layers. This is a reasonable assumption due to the fact that the training data used is far from sufficient compared to the network parameters.

The training data is prepared as following. First, we extract top 1000 object proposals within each image. The proposal boxes are then augmented with random shifting or magnifying/reducing along two dimensions by maximum 20\%, resulting in additional 1000 object proposals for each image. Second, for all generated proposals, their ground-truth scores are calculated according to Section~\ref{subsec:quality_assessment_criteria}. Third, the proposals within each image are randomly sampled to be used in network training. Note that the scores may concentrate around certain regions, depending on the performance of specific proposal generators. For example, proposal windows produced by better generators may contain more boxes with ground-truth objectness scores over 0.5 than those scored below 0.5. This requires us to do sampling carefully such that the training data is evenly distributed in different ground-truth score ranges. In this work, we sample proposals according to their respective ground-truth score ranges. For each image, we calculate the histogram of ground-truth scores over all proposals, then randomly sample boxes within each histogram bin to maintain a generally even distribution of scores for training. The training images are randomly divided into 5 disjoint sets, one of which is used as validation set during training.

During the training stage, each mini-batch contains one image and at most 256 proposal windows. For the sake of correctness and efficiency of training, the proposal windows included in a mini-batch are always generated from the same image. Gradients of layer weights with respect to the training loss are calculated using finite difference and backward propagation, while weights updating is through stochastic gradient descent. In addition, we use a batch normalization layers~\cite{batchnorm-2015-ioffe} for larger base learning rate and faster convergence. The initial learning rate is set to 0.001, momentum parameter is 0.9 and the weight decay is 0.0005. We test the model performance with respect to validation set at the end of each epoch. After each epoch, the order of images and proposal boxes are randomly shuffled. A k-fold cross-validation with $k=5$ is used here. The configuration of our deep-learning server: Ubuntu 14.04 64-bit (OS), Intel Xeon E5-2699-v3 2.30GHz (64 CPUs), Nvidia GTX Titan X 6GB (1 GPU), 128GB (RAM).


\section{Experimental Results}\label{sec:evaluation}

\subsection{Datasets and Evaluation Criteria}

We evaluate our method on three public datasets: PASCAL VOC 2007 \cite{voc_2007}, PASCAL VOC 2012 \cite{voc_2012} and SOS \cite{sod-cvpr2015-zhang}. PASCAL VOC 2007 and PASCAL VOC 2012 contain 9963 and 11540 images respectively with fully annotated bounding boxes for 20 object categories. As we focus on class-agnostic object locating, we only use bounding box locations in these datasets for training and evaluation. To conduct fair comparison, we train our network on the same training data that is used by the other object proposal generators. To evaluate the generalization performance of our trained networks, we also test their performance on the SOS dataset, a group of 5244 images hand-picked from COCO \cite{coco-cvpr2014-lin}, ImageNet \cite{imagenet-ijcv2015-russakovsky}, SUN \cite{sun-cvpr2010-xiao}, and PASCAL VOC \cite{voc_2007, voc_2012}. 3951 images in the SOS dataset have been annotated with class-agnostic object windows. Since we train our networks on PASCAL VOC datasets, we remove images in SOS dataset that come from VOC datasets. In the end, 3086 annotated images are left in the SOS dataset for our experiments.

When we compare our method with other state-of-the-art proposal generators, we adopt three criteria, {\em precision rate}, {\em recall rate} and {\em mean ground-truth score}. The precision rate is the percentage of detected objects among all object proposals, while the recall rate is the percentage of detected objects among all ground-truth objects. One important factor in calculating precision and recall is to decide whether or not an object proposal is selected. In our experiments, a proposal is selected to be any proposal window with intersection-over-union (IOU) between it and a ground-truth box reaches 0.7. Note that this is relatively a high IOU threshold, compared to what is commonly used in object detection applications. In this work, precision and recall are discussed separately with respect to different numbers of top proposals. For some models trained with specific settings, the performance of area-under-curve (AUC) is also reported. In addition, we adopt the mean ground-truth score as the third criterion, which also reflects the general quality of top-ranked proposals. The ground-truth score of a proposal window is calculated using methods described in Section~\ref{subsec:quality_assessment_criteria}.

\subsection{Optimizing Ground-truth Scoring Schemes}\label{subsec:scheme_optimize}

In this work, The weighting parameter $w$ in Equation~\ref{eq:final_quality_score} is decided according to the network performance on the validation set of PASCAL VOC2012. We use the same network structure in Section~\ref{sec:network_architecture} and the same training parameters in Section~\ref{sec:network_training}. The $w$ is selected such that the network gives the best precision on validation data. In our experiments, several $w$ values are tested. Table~\ref{tab:aucs_w} shows the performance comparison on testing data among models trained with different weights. In our experiments, the optimal $w$ turns out to be dependent on proposal generators (Table~\ref{tab:aucs_w}). In the rest of the paper, we report the performance of the network trained on proposals generated by COB, unless specified otherwise.

We also try linear transfer functions for calculating {\em completeness score} and {\em fullness score}, namely, $S_{linear} = w C_{c} + (1-w) C_{f}$, as plotted by grey dashed lines in Figure~\ref{fig:calc_completeness_fullness_score}. The results are in Table~\ref{tab:aucs_schemes} (``Linear''). On all testing data (VOC2012, VOC2007, SOS), the precision and recall of the network trained using linear transfer functions are generally worse than those trained with non-linear transfer functions described in Section~\ref{subsec:quality_assessment_criteria}, by approximately 1\%-3\% regarding precision and by up to 3\% regarding recall. The reason should be that the non-linear transfer functions used in this work enlarge the gap between the scores of high-objectness and low-objectness proposals, giving rise to improved precision and recall when top-ranked proposal windows within each image are considered. The using of non-linear transfer functions also offers flexibility in dealing with {\em completeness} and {\em fullness} differently, such as tolerating insignificant violation of {\em fullness} (using generalized logistic function) and penalizing any violation of {\em completeness} (using quadratic function).

Intuitively, instead of using methods described in Section~\ref{subsec:quality_assessment_criteria}, a straightforward idea of calculating ground-truth scores is to directly use IOU between proposal box and ground-truth box in network training. However, in our experiments, we find that using IOU instead of our proposed method leads to up to 3\% worse precision and up to 2\% worse recall, as suggested by Table~\ref{tab:aucs_schemes}. The reason should be that IOU only captures limited information (i.e., the percentage of overlap area between two boxes), but does not reflect the positional relationship between them.

\begin{figure*}
	\centering
	\captionsetup{justification=centering}
	
	\begin{subfigure}[]{0.32\linewidth}
		\includegraphics[width=\linewidth]{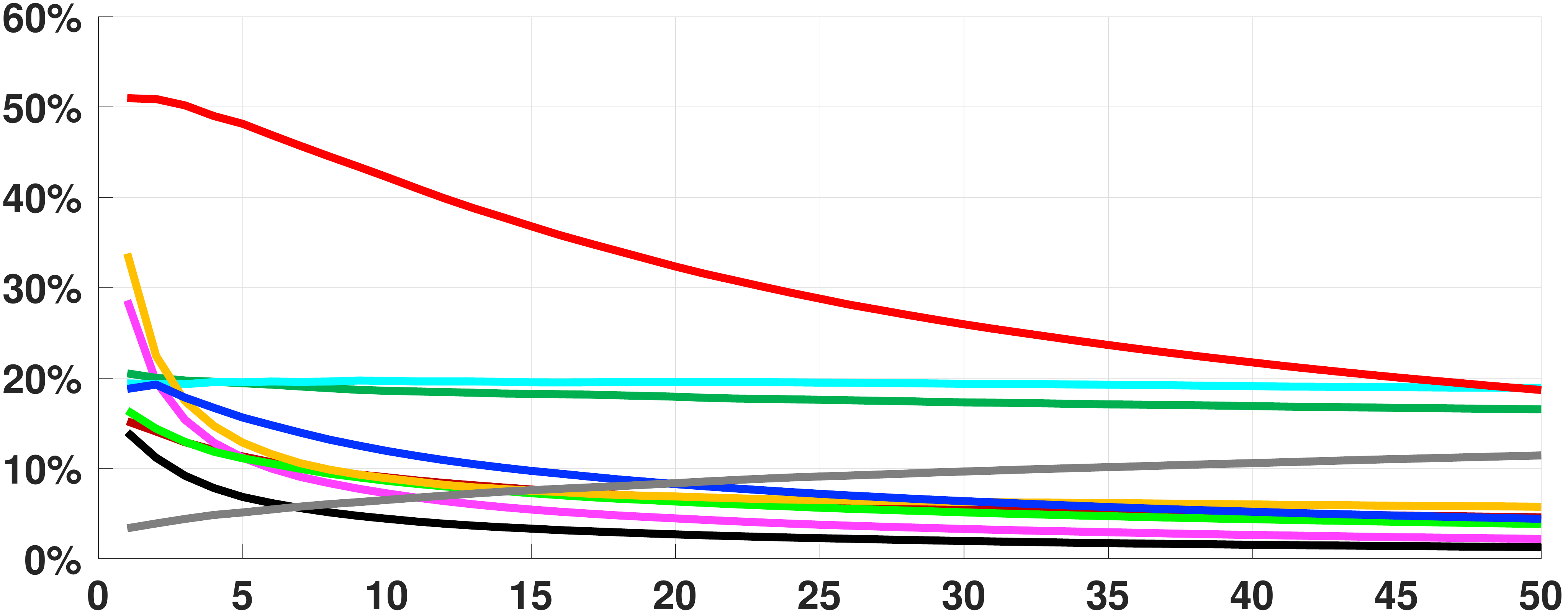}
		\caption{Precision on VOC2012}
		\label{fig:06_01_voc2012_precision}
	\end{subfigure}
		\begin{subfigure}[]{0.32\linewidth}
		\includegraphics[width=\linewidth]{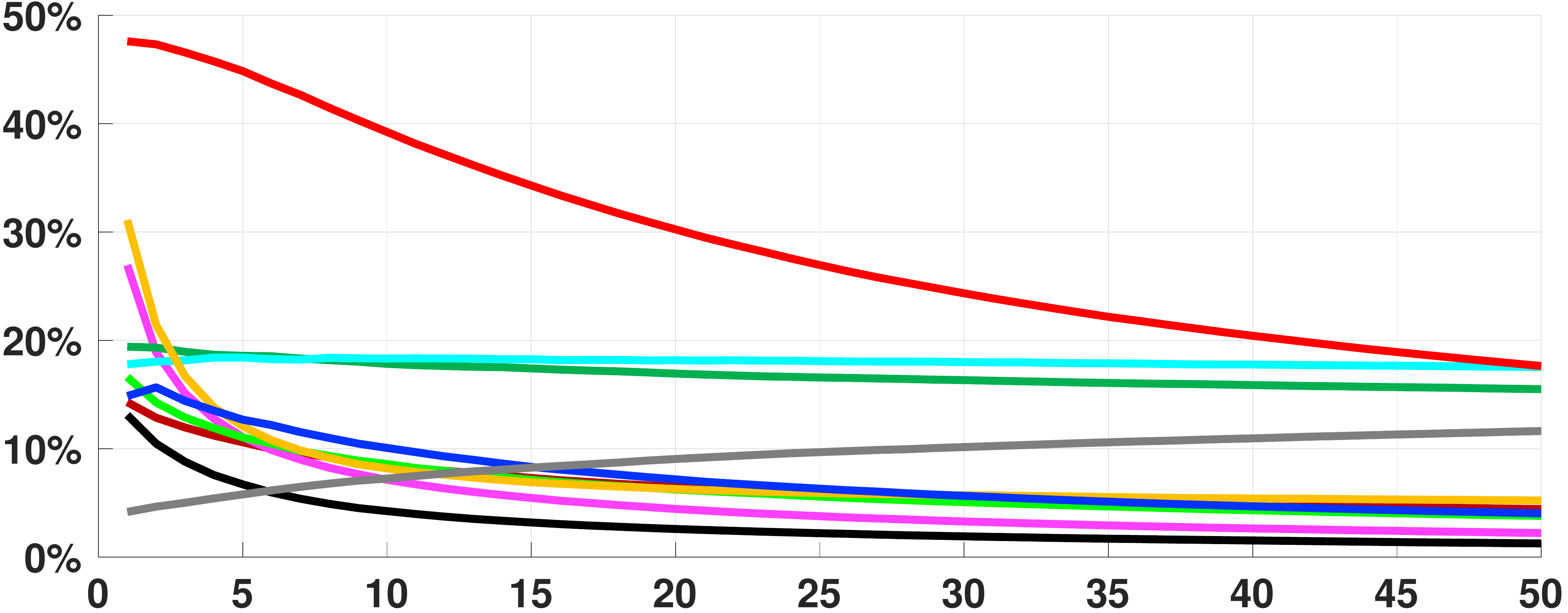}
		\caption{Precision on VOC2007}
		\label{fig:06_01_voc2007_precision}
	\end{subfigure}
	\begin{subfigure}[]{0.32\linewidth}
		\includegraphics[width=\linewidth]{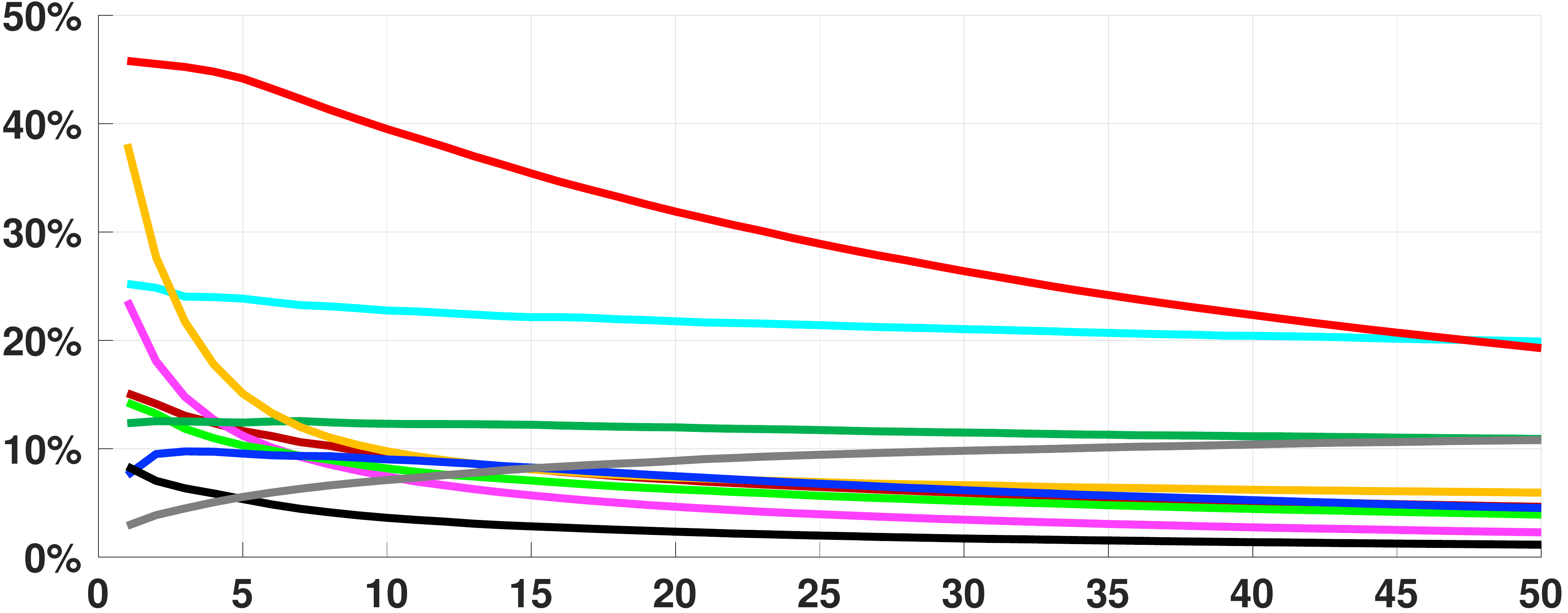}
		\caption{Precision on SOS}
		\label{fig:06_01_sos_precision}
	\end{subfigure}
	
	\vspace{5pt}
	
	\begin{subfigure}[]{0.32\linewidth}
		\includegraphics[width=\linewidth]{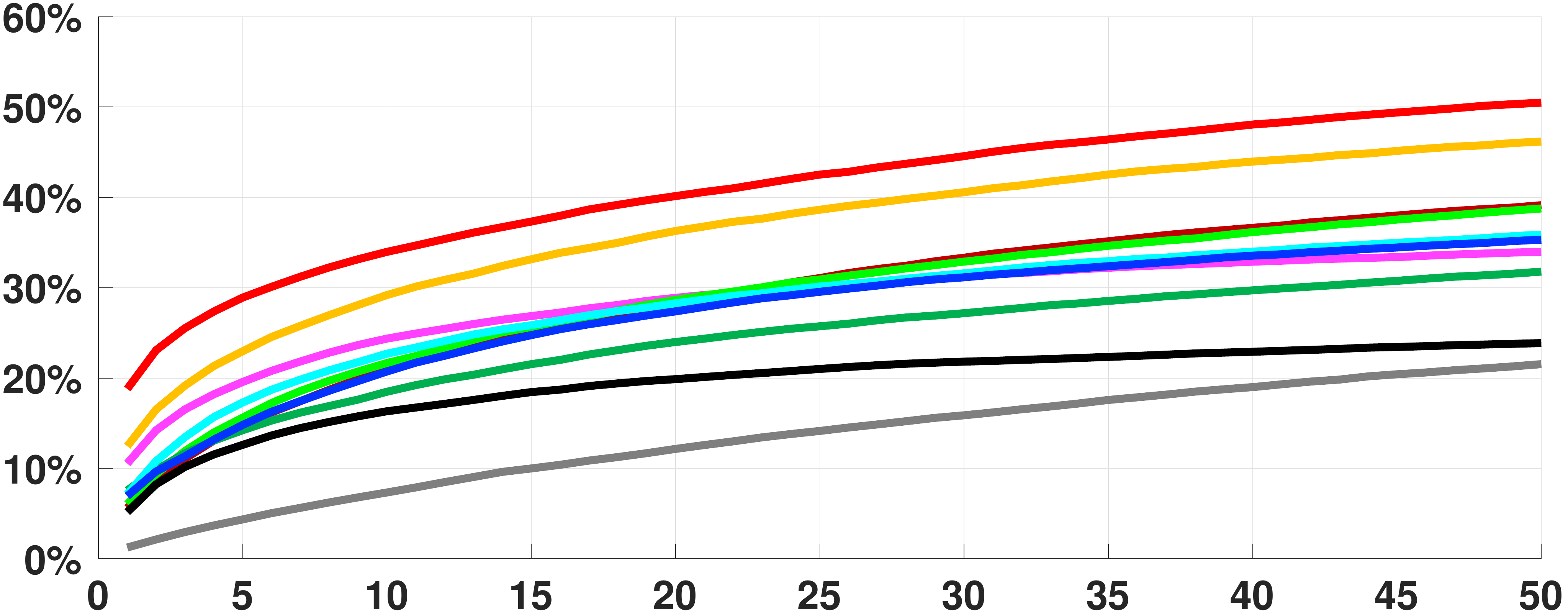}
		\caption{Recall on VOC2012}
		\label{fig:06_01_voc2012_recall}
	\end{subfigure}
	\begin{subfigure}[]{0.32\linewidth}
		\includegraphics[width=\linewidth]{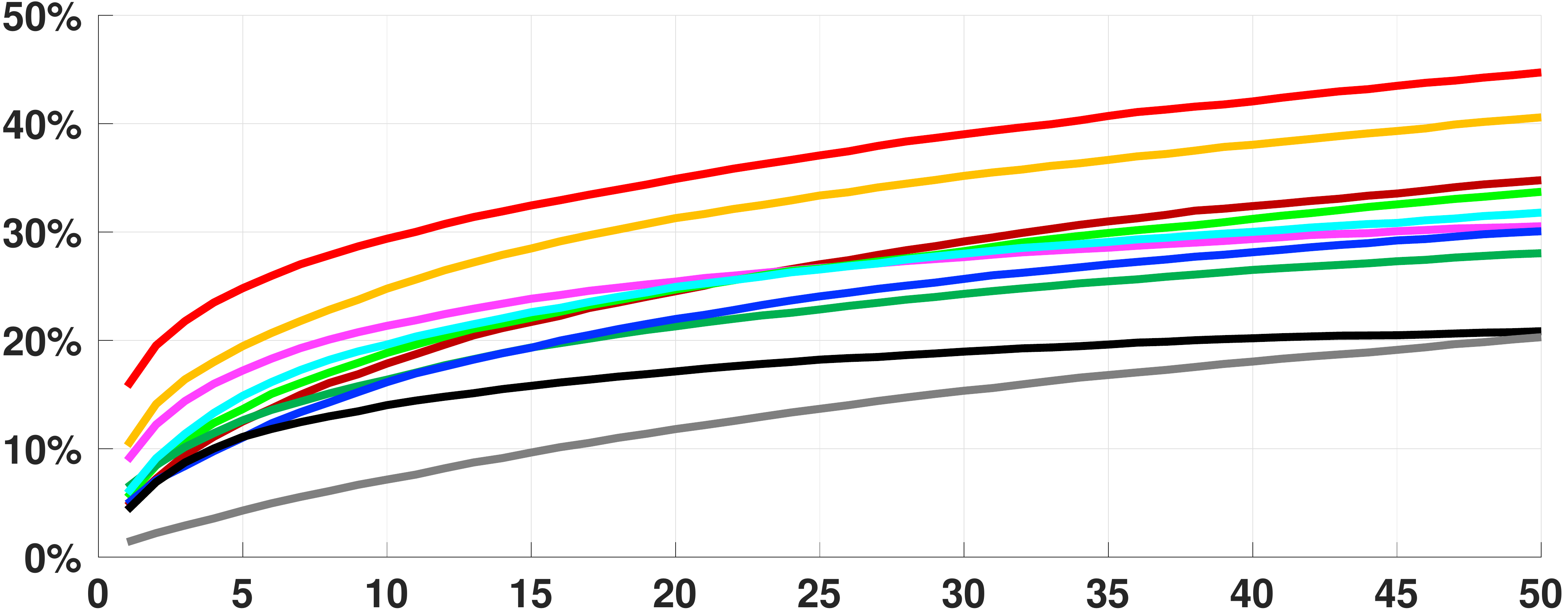}
		\caption{Recall on VOC2007}
		\label{fig:06_01_voc2007_recall}
	\end{subfigure}
	\begin{subfigure}[]{0.32\linewidth}
		\includegraphics[width=\linewidth]{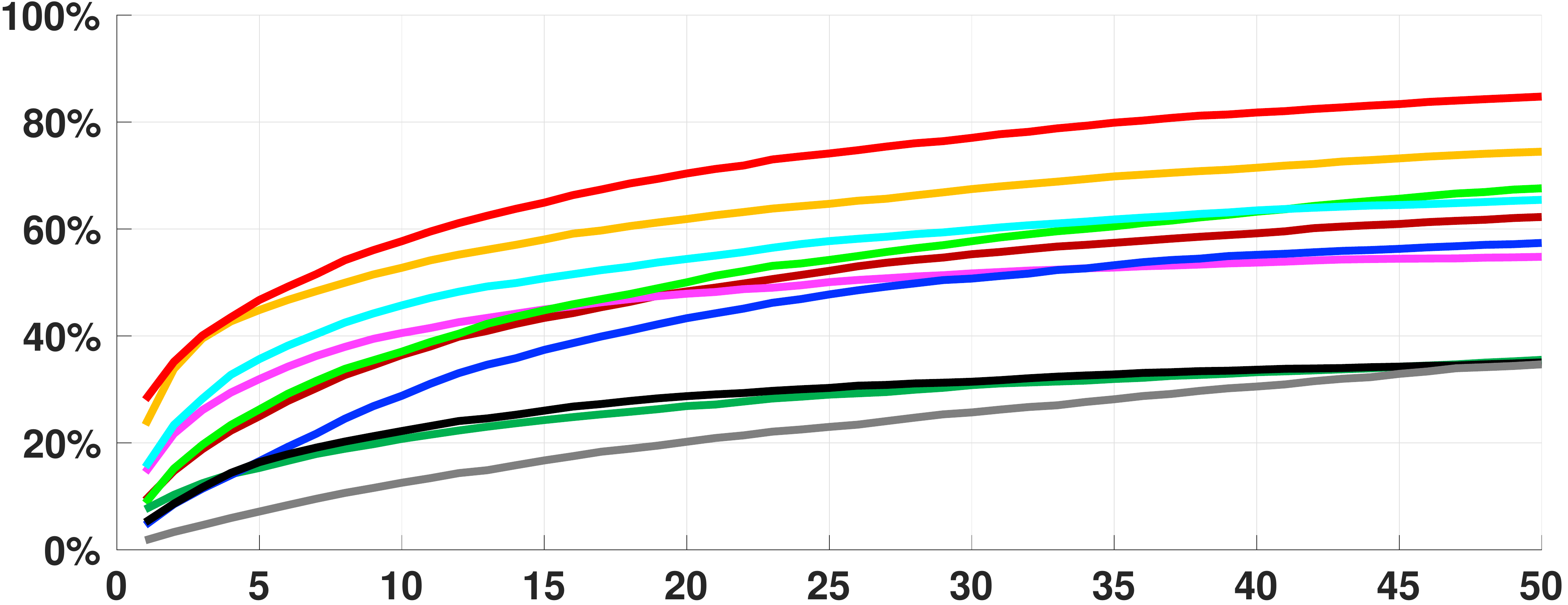}
		\caption{Recall on SOS}
		\label{fig:06_01_sos_recall}
	\end{subfigure}
	
	\vspace{8pt}
	
	\begin{subfigure}[]{0.85\linewidth}
		\includegraphics[width=\linewidth]{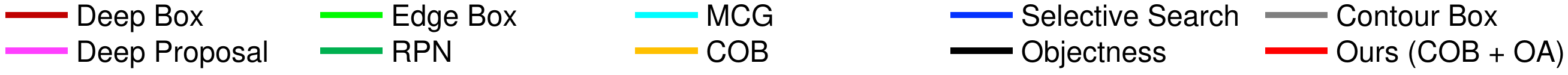}
	\end{subfigure}
	
	\caption{Performance comparison between different object proposal methods. Horizontal axis: number of top-ranked proposals.}
	\label{fig:06_01_comparison}
	
\end{figure*}

\subsection{Comparison with State of the Art Methods}\label{subsec:comparison_external}

Our proposal objectness assessment network takes an image and its object proposals as input and returns objectness scores, which are used to re-rank the original object proposals. We have tried several proposal methods in building the network. In the rest of the paper, each proposal objectness assessment network is denoted as ``<Proposal Generator Name>+OA''. For example, network trained on COB proposals is denoted as ``COB+OA''. We compare COB+OA against nine state-of-the-art object proposal generators, including Objectness Measure~\cite{alexe2012measuring}, Contour Box~\cite{contour_box-iccv2015-lu}, Selective Search~\cite{selective_search-ijcv2013-uijlings}, Edge Box~\cite{edgebox-eccv2014-zitnick}, MCG~\cite{mcg-cvpr2014-arbelaez}, COB~\cite{cob-eccv2016-maninis}, Deep Box~\cite{deep_box-iccv2015-kuo}, Deep Proposal~\cite{deep_proposal-iccv2015-ghodrati}, and RPN~\cite{faster_rcnn-nips15-ren}. The last four are the latest deep learning based methods. We use their publicly released models and suggested parameters for the reproduction of their best results. We evaluate the precision and recall rates over three datasets: the validation set of PASCAL VOC 2012, the testing set of PASCAL VOC 2007 and our customized SOS dataset.

As shown in Figure~\ref{fig:06_01_comparison}, when at most 50 top-ranked proposal windows per image are considered, our method (COB+OA) significantly outperforms all other participating methods across all testing datasets in terms of both precision and recall. Specifically, Figures~\ref{fig:06_01_voc2012_precision},~\ref{fig:06_01_voc2007_precision} and~\ref{fig:06_01_sos_precision} respectively report precision with respect to the number of top proposals. The AUCs (Areas Under Curve) of our hybrid method (COB+OA) are 31.15\%, 29.12\% and 30.34\%, which improve those achieved by the best-performing existing algorithm by 11.78\%, 11.11\% and 8.72\%, respectively, on PASCAL VOC 2012, PASCAL VOC 2007 and our customized SOS dataset. When the number of considered proposal windows further increases, the advantage of our method in terms of precision rate diminishes. However, in our case, each image can hardly contain too many useful objects. Therefore, with a proper non-maximal suppression threshold, checking only highly-ranked proposals is enough for extracting good objects. The precision of our method also confirms that a significant improvement can be achieved through re-ranking existing object proposal windows. As for the recall rate, as shown in Figures \ref{fig:06_01_voc2012_recall}, \ref{fig:06_01_voc2007_recall} and \ref{fig:06_01_sos_recall}, our method also noticeably outperforms all other participating methods. The AUCs of our method (COB+OA) are 4.42\%, 4.21\% and 10.31\% higher than those achieved by the best-performing existing algorithm, respectively, on PASCAL VOC 2012, PASCAL VOC 2007 and the screened SOS dataset. This implies that our method can include more ground-truth objects into the database than other methods when highly-ranked proposals are considered.

\begin{figure*}
	\centering
	\captionsetup{justification=centering}
	
	\begin{subfigure}[]{0.32\linewidth}
		\includegraphics[width=\linewidth]{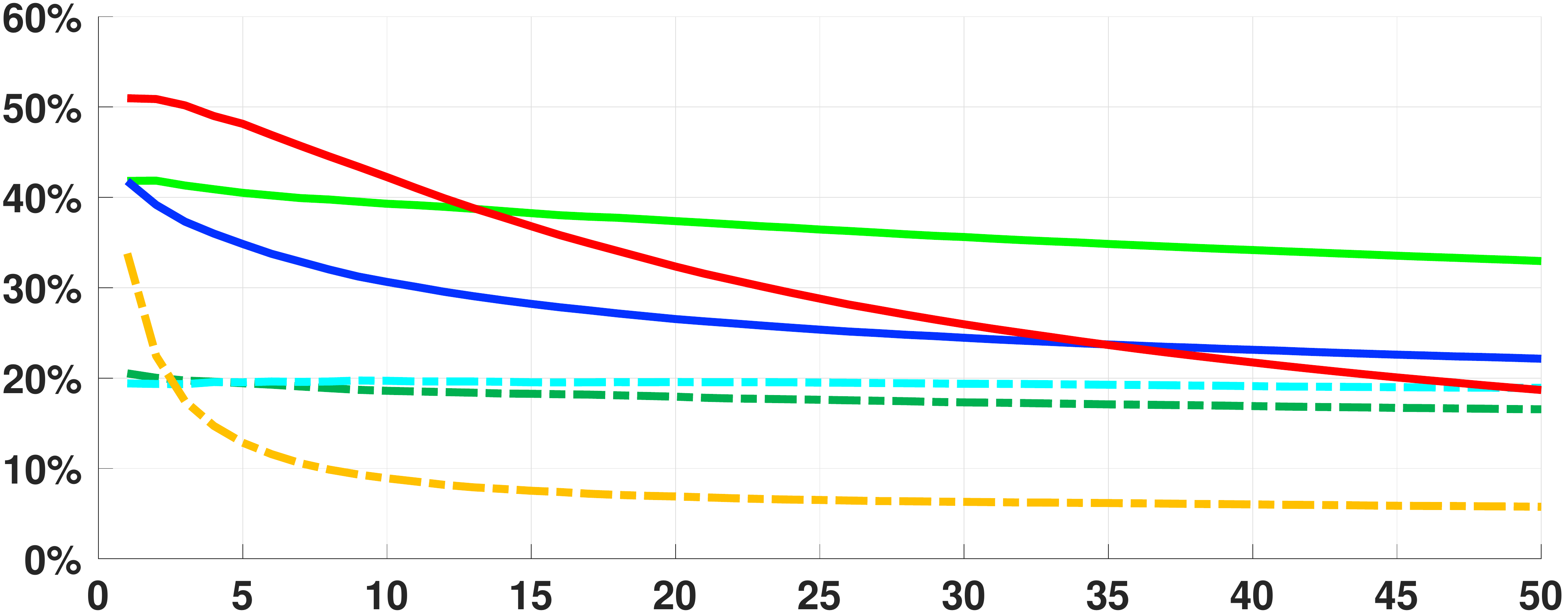}
		\caption{Precision on VOC2012}
		\label{fig:06_02_voc2012_precision}
	\end{subfigure}
	\begin{subfigure}[]{0.32\linewidth}
		\includegraphics[width=\linewidth]{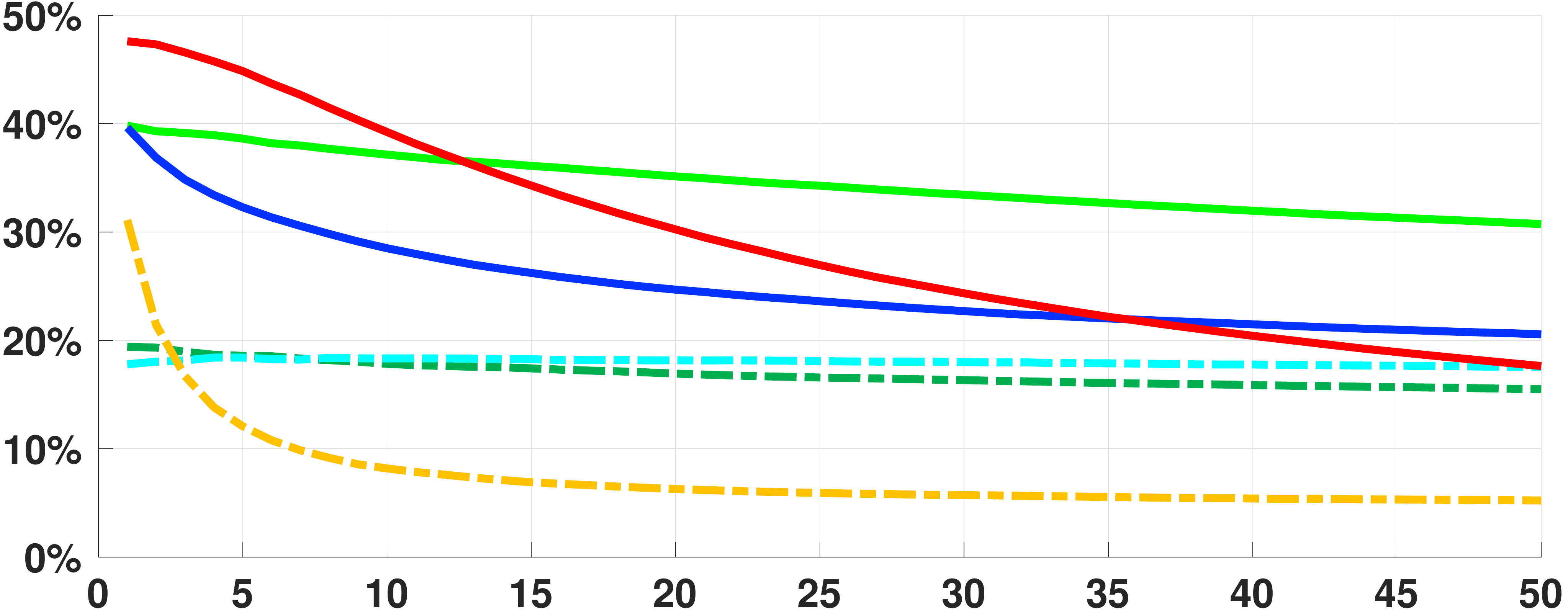}
		\caption{Precision on VOC2007}
		\label{fig:06_02_voc2007_precision}
	\end{subfigure}
	\begin{subfigure}[]{0.32\linewidth}
		\includegraphics[width=\linewidth]{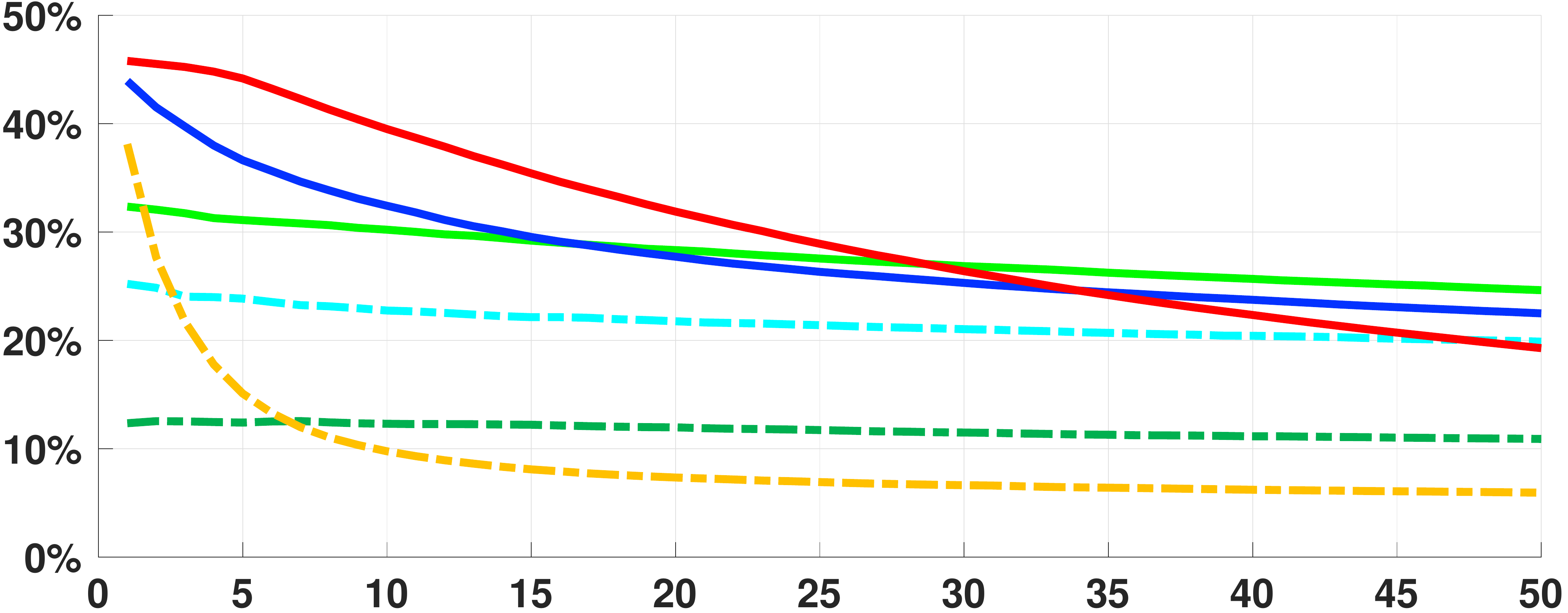}
		\caption{Precision on SOS}
		\label{fig:06_02_sos_precision}
	\end{subfigure}
	
	\vspace{5pt}
	
	\begin{subfigure}[]{0.32\linewidth}
		\includegraphics[width=\linewidth]{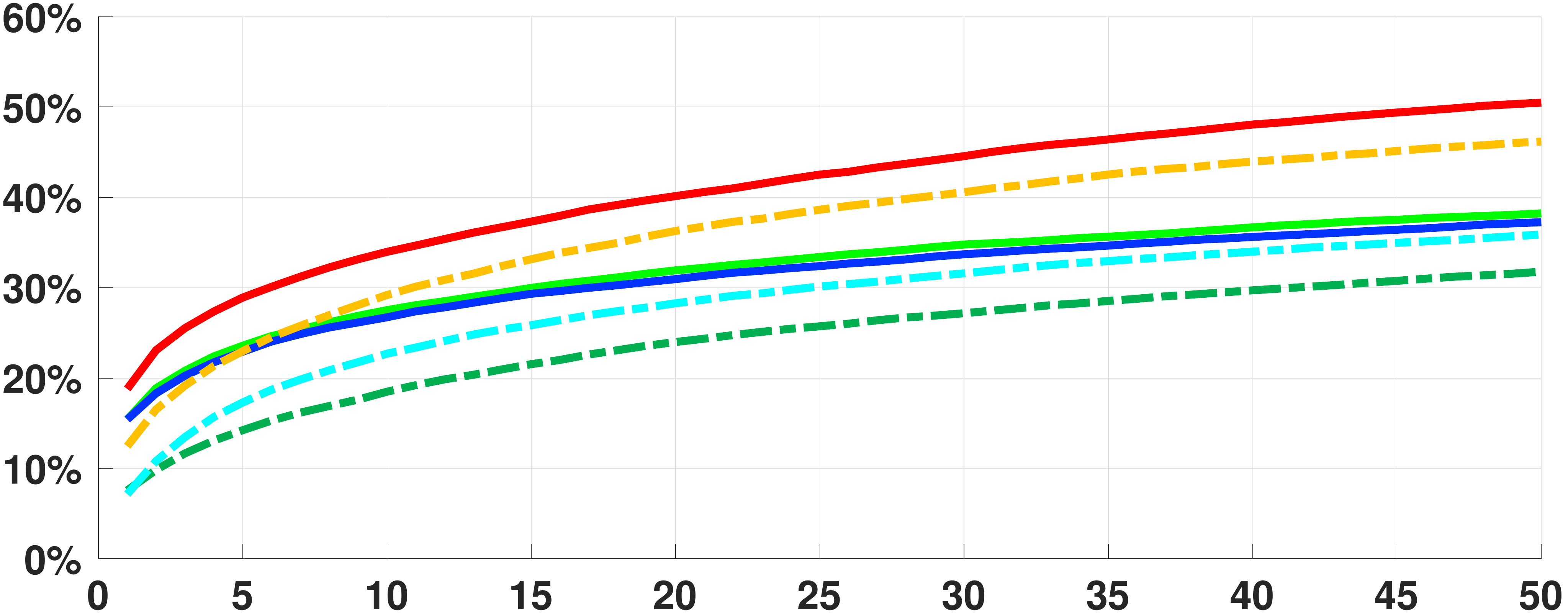}
		\caption{Recall on VOC2012}
		\label{fig:06_02_voc2012_recall}
	\end{subfigure}
	\begin{subfigure}[]{0.32\linewidth}
		\includegraphics[width=\linewidth]{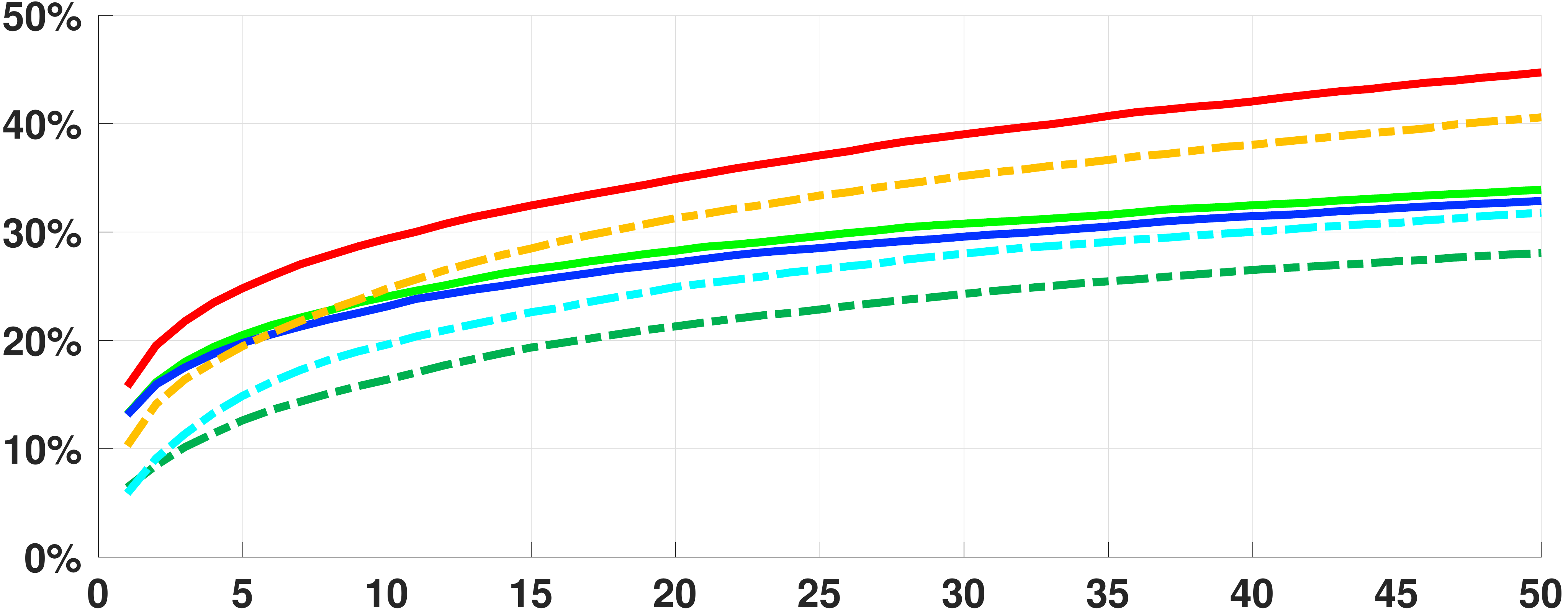}
		\caption{Recall on VOC2007}
		\label{fig:06_02_voc2007_recall}
	\end{subfigure}
	\begin{subfigure}[]{0.32\linewidth}
		\includegraphics[width=\linewidth]{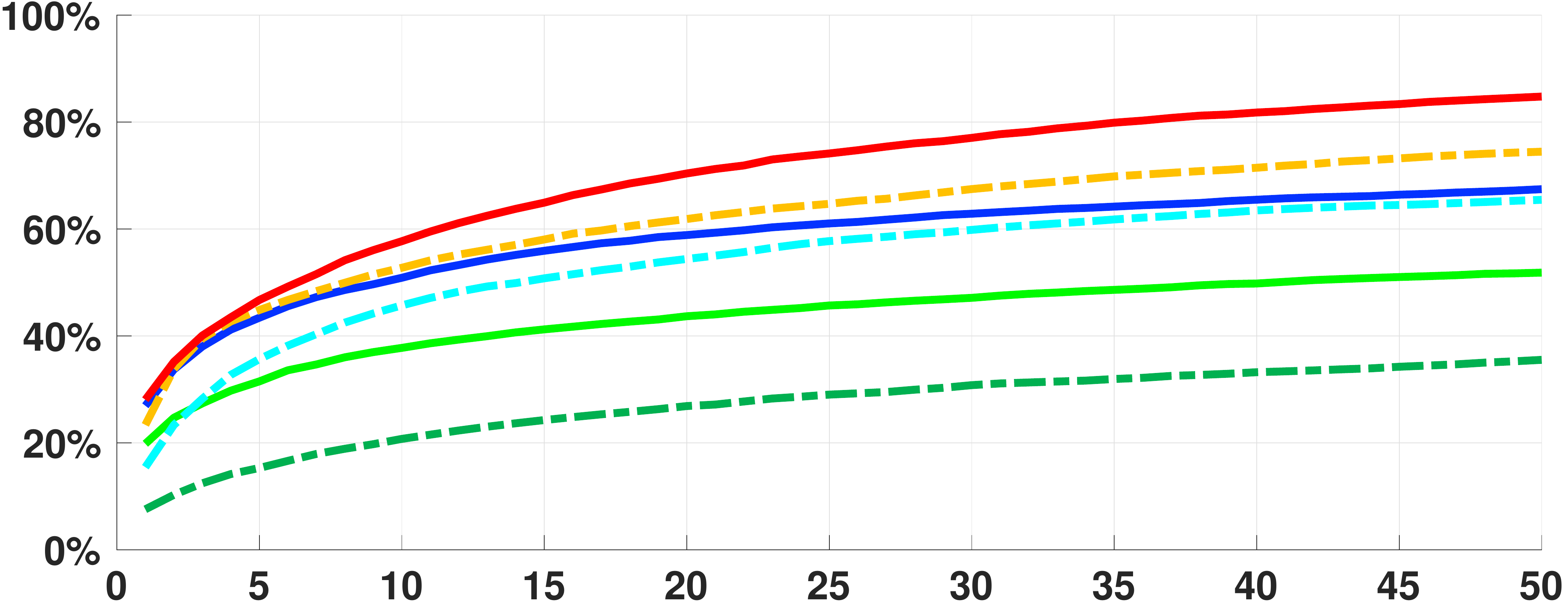}
		\caption{Recall on SOS}
		\label{fig:06_02_sos_recall}
	\end{subfigure}
	
	\vspace{5pt}
	
		\begin{subfigure}[]{0.32\linewidth}
		\includegraphics[width=\linewidth]{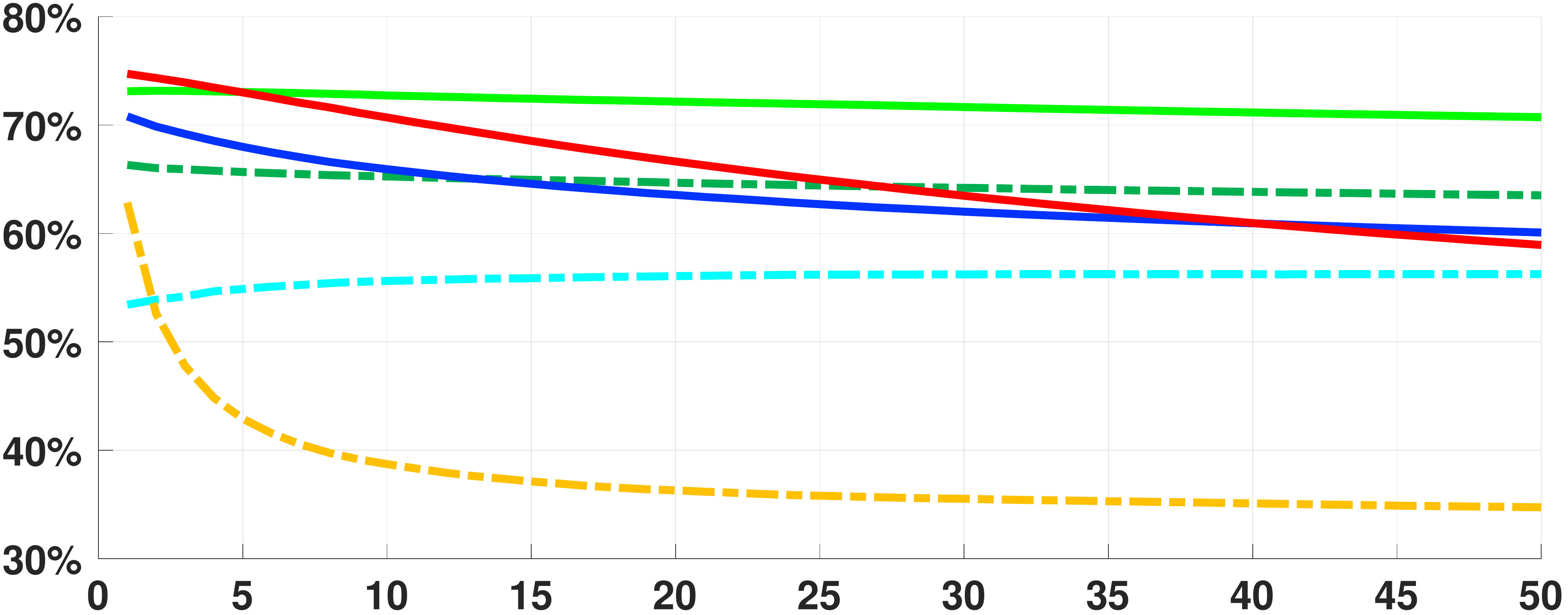}
		\caption{Mean score on VOC2012}
		\label{fig:06_02_voc2012_avgmark}
	\end{subfigure}
	\begin{subfigure}[]{0.32\linewidth}
		\includegraphics[width=\linewidth]{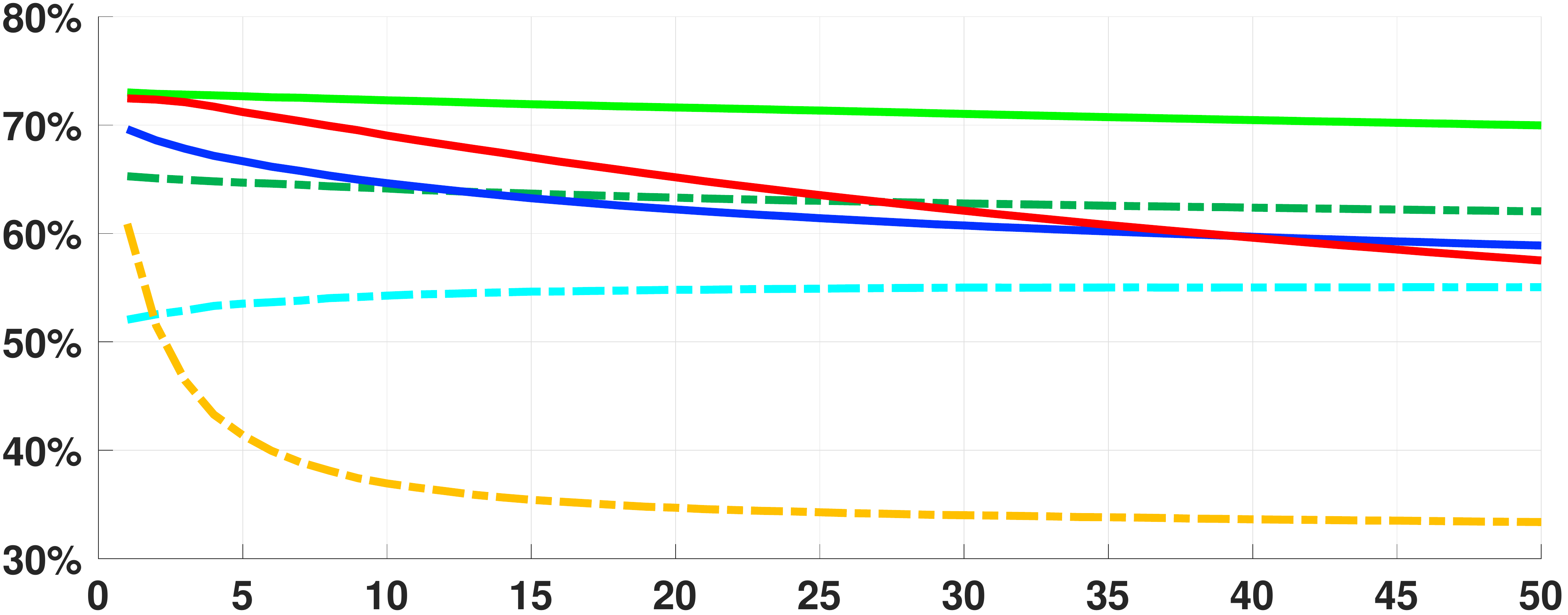}
		\caption{Mean score on VOC2007}
		\label{fig:06_02_voc2007_avgmark}
	\end{subfigure}
	\begin{subfigure}[]{0.32\linewidth}
		\includegraphics[width=\linewidth]{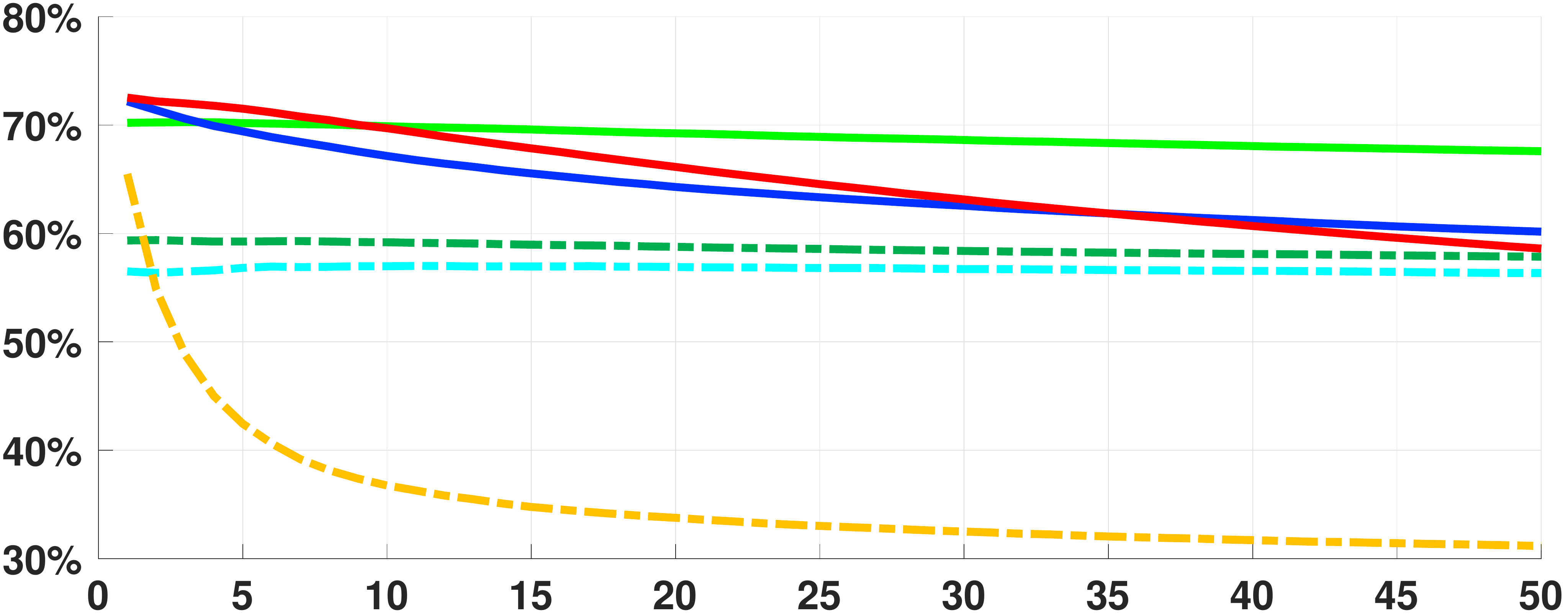}
		\caption{Mean score on SOS}
		\label{fig:06_02_sos_avgmark}
	\end{subfigure}
	
	\vspace{8pt}
	
	\begin{subfigure}[]{0.40\linewidth}
		\includegraphics[width=\linewidth]{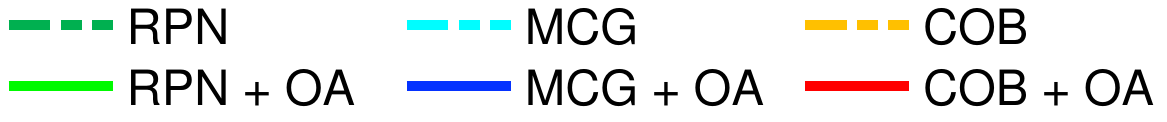}
	\end{subfigure}
	
	\caption{Performance comparison between models with or without the integration of objectness assessment network. Horizontal axis: number of top-ranked proposals.}

	\label{fig:06_02_comparison}
	
\end{figure*}

\subsection{General Effectiveness of Proposal Objectness Assessment}

In this section, we verify that our network has performance gains on top of most object proposal methods. To this end, we check the performance of a few object proposal methods and their respective assessment networks. For each of the object proposal methods, we train a new objectness assessment network using proposal windows generated by that method. To conduct a fair comparison, only the original training data used for an object proposal method is used to train its associated network. For example, if the image set ``trainval'' in PASCAL VOC 2007 is used in training MCG, we only use the same image set in training our corresponding network. All the networks are trained using the same settings as described in Section~\ref{sec:network_training}.

We evaluate the effectiveness of our method on three state-of-the-art object proposal methods, RPN~\cite{faster_rcnn-nips15-ren}, MCG~\cite{mcg-cvpr2014-arbelaez} and COB~\cite{cob-eccv2016-maninis}, and report the precision, recall and mean ground-truth score on the same testing set used in Section~\ref{subsec:comparison_external}. As shown in Figure~\ref{fig:06_02_voc2012_precision},~\ref{fig:06_02_voc2007_precision}, and~\ref{fig:06_02_sos_precision}, our method significantly improves the precision of all three original object proposal methods. Specifically, as reported in Table~\ref{tab:aucs_score}, our method boosts AUCs of the original COB, MCG and RPN by 22.90\%, 7.57\% and 18.88\%, respectively, on the validation set of PASCAL VOC 2012. Meanwhile, as shown in Figures~\ref{fig:06_02_voc2012_recall}, \ref{fig:06_02_voc2007_recall} and \ref{fig:06_02_sos_recall}, our method also improves the recall rate of all three original object proposal methods, increasing their AUCs by 4.42\%, 2.92\% and 12.28\%, respectively, on PASCAL VOC 2012. We can observe similar performance gains on the other two datasets from Table~\ref{tab:aucs_score}.

Figure~\ref{fig:06_02_voc2012_avgmark},~\ref{fig:06_02_voc2007_avgmark}, and~\ref{fig:06_02_sos_avgmark} show the mean ground-truth scores evaluated on different methods. Our method significantly improves the mean ground-truth scores of all three original object proposal methods using our own standards. These results confirm that our objectness assessment network can be used as a generic tool for improving the accuracy of object proposal methods.

\begin{figure*}
	\centering
	
	\begin{subfigure}[]{0.32\linewidth}
		\includegraphics[width=\linewidth]{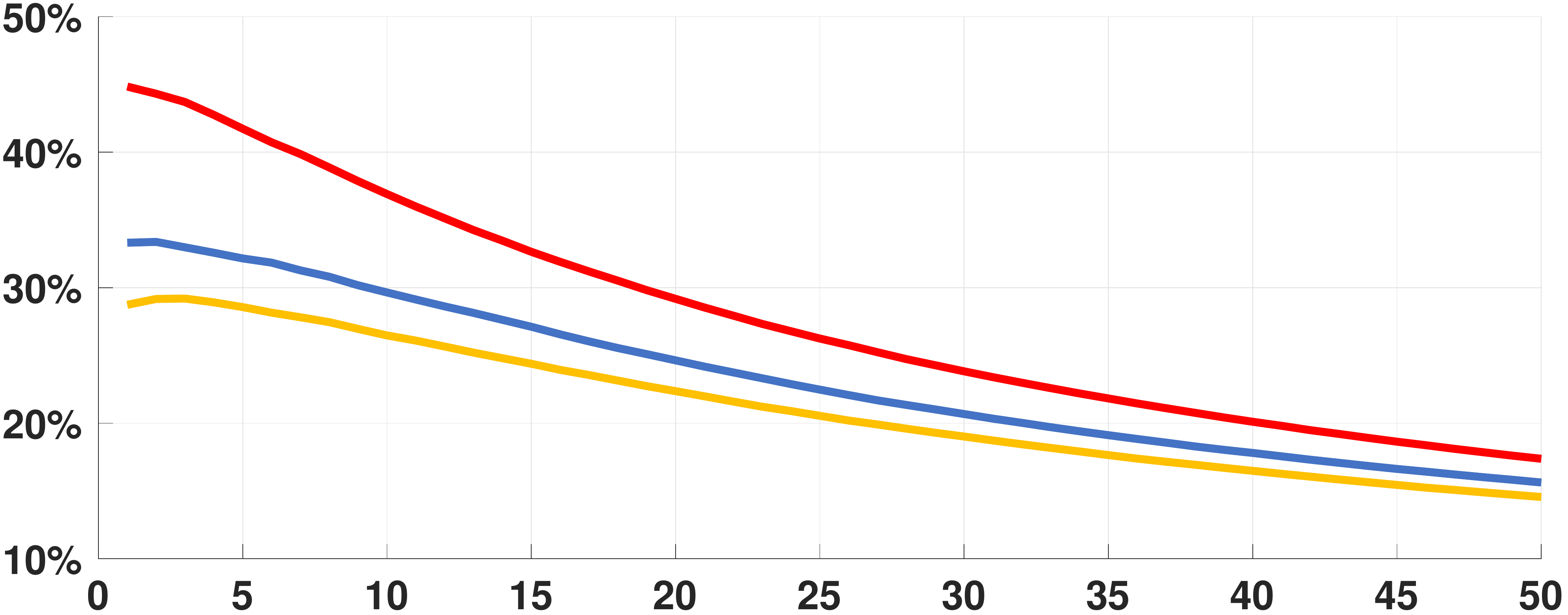}
		\caption{Precision on VOC2012}
		\label{fig:06_04_voc2012_precision}
	\end{subfigure}
	\begin{subfigure}[]{0.32\linewidth}
		\includegraphics[width=\linewidth]{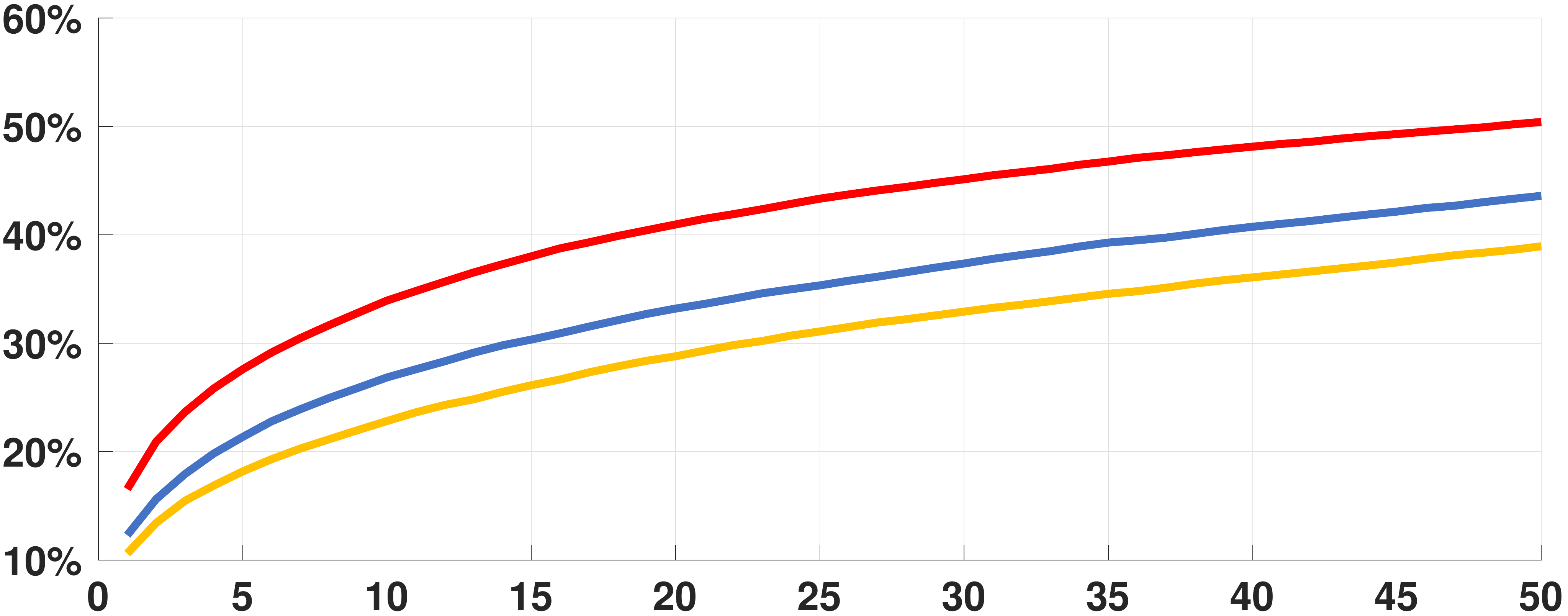}
		\caption{Recall on VOC2012}
		\label{fig:06_04_voc2012_recall}
	\end{subfigure}
	\begin{subfigure}[]{0.32\linewidth}
		\includegraphics[width=\linewidth]{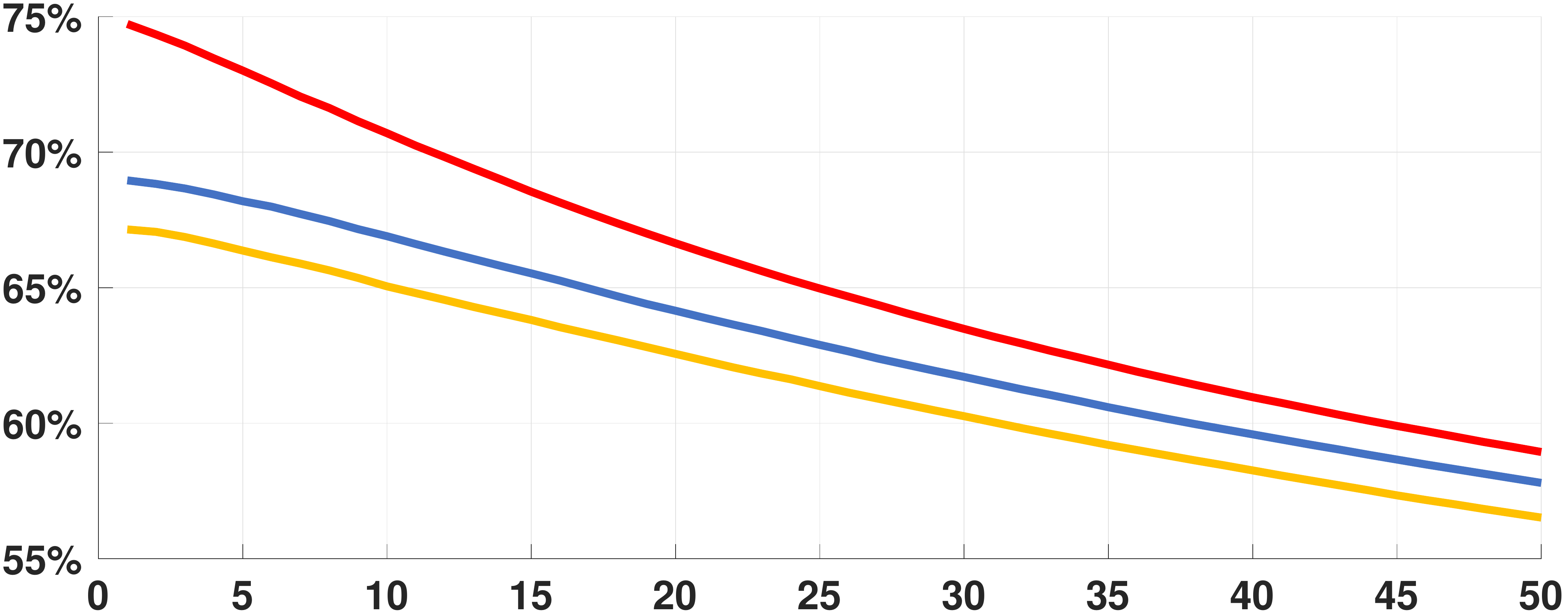}
		\caption{Mean score on VOC2012}
		\label{fig:06_04_voc2012_avgmark}
	\end{subfigure}
	
	\vspace{8pt}
	
	\begin{subfigure}[]{0.9\textwidth}
		\includegraphics[width=\linewidth]{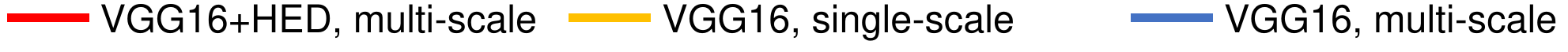}
	\end{subfigure}
	
	\caption{Performance comparison between networks with different architectures. Horizontal axis: number of top-ranked proposals. ``VGG16+HED'': the proposal objectness assessment network consists of both VGG16~\cite{vgg16-corr2014-simonyan} and HED~\cite{hed-iccv2015-xie} as sub-networks for feature extraction. ``VGG16'': the proposal objectness assessment only utilizes VGG16~\cite{vgg16-corr2014-simonyan} for feature extraction. ``multi-scale'': ROI pooling is performed on three different feature maps ``conv\_3-3'', ``conv\_4-3'', ``conv\_5-3''. ``single-scale'': ROI pooling is only performed on one feature map ``conv\_5-5''.}

	\label{fig:06_04_design_comparison}
	
\end{figure*}

\subsection{Ablation Study of Network Architecture}

An ablation study is conducted to verify the effectiveness of the proposed network architecture. We replace the two parallel sub-networks in our architecture with simpler versions. One of them has the same convolutional layers as Faster R-CNN~\cite{faster_rcnn-nips15-ren}, but has only one sub-network for feature extraction. The ROI pooling is only applied on ``conv5-3" layer, further followed by fully connected layers. The other simpler version does not have the second sub-network for edge-oriented feature extraction but keeps multi-scale ROI pooling. The fully connected layers receive concatenated features from the ROI pooling layers associated with ``conv3-3'', ``conv4-3'' and ``conv5-3'' respectively. These three networks are trained on the PASCAL VOC 2012 training set using the same settings described in Section~\ref{sec:network_training}. The testing results on PASCAL VOC 2012 are shown in Figure~\ref{fig:06_04_design_comparison}, which includes precision (Figure \ref{fig:06_04_voc2012_precision}), recall (Figure \ref{fig:06_04_voc2012_recall}) and the mean ground-truth score. It is clearly shown that our network architecture delivers better performance than the two simpler alternatives.


\subsection{Number of Object Proposals Generated during Training}

We investigate how the number of proposals generated during the training stage can affect the performance of assessment network. In data preparation, we empirically generate 1000 proposal windows per image. However, this number can vary in practical cases, depending on the fact that how many proposals are enough for achieving good precision and recall. As it is obvious that less generated proposals mean better efficiency of the whole pipeline. However, reducing generated proposals also reduces the richness of training data. Although we have data augmentation, the richness of training data can only be improved to a certain extent. Table~\ref{tab:bboxN} shows the performance when different numbers of proposal windows (per image) are generated during training stages. It can be seen in Table~\ref{tab:bboxN} that changing the number of bounding boxes from 1000 to 500 does not negatively affect the performance much (approximately within 1\% difference regarding precision and recall). However, if the generated proposals are too few, the performance can decrease significantly, with the precision dropping by around 2\%-3\%. In our experiments, increasing the number of proposals from 1000 to 2000 does not significantly affect the performance, with precision and recall within 1\% difference.

\begin{table*}
	\centering
	\caption{Comparison of AUCs of precision and recall curves between models trained with different scoring schemes. \textbf{Ours}: scoring scheme described in Section~\ref{subsec:quality_assessment_criteria}. \textbf{Linear}: scoring scheme with linear transfer functions (grey dash-lines in Figure~\ref{fig:calc_completeness_fullness_score}). \textbf{IOU}: Intersection-over-union as proposal objectness scores.}
	\label{tab:aucs_schemes}
	\resizebox{0.5\textwidth}{!}
	{
		\begin{tabu}{|[1.5pt]c|[1.5pt]c|c|c|[1.5pt]c|c|c|[1.5pt]}

		\tabucline[1.5pt]{-}

		\multirow{2}{*}{\textbf{Dataset}} & \multicolumn{3}{c|[1.5pt]}{\textbf{Precision Curve AUCs}} & \multicolumn{3}{c|[1.5pt]}{\textbf{Recall Curve AUCs}}\\

		\tabucline[1pt]{2-7}

		\textbf{} & \textbf{IOU} & \textbf{Linear} & \textbf{Ours} & \textbf{IOU} & \textbf{Linear} & \textbf{Ours}\\

		\tabucline[1.5pt]{-}

		\textbf{VOC2012} & 29.11\% & 27.98\% & 31.15\% & 38.10\% & 38.75\% & 40.94\%\\ \hline
		\textbf{VOC2007} & 26.73\% & 28.41\% & 29.12\% & 33.28\% & 32.38\% & 35.73\%\\ \hline
		\textbf{SOS} & 29.28\% & 28.74\% & 30.34\% & 68.75\% & 69.65\% & 69.87\%\\ \hline

		\tabucline[1.5pt]{-}

		\end{tabu}
	}
\end{table*}

\begin{table*}[]
	\centering
	\caption{Comparison of AUCs of precision curves between models trained with different weights. \textbf{${w}$}: weight used in Equation~\ref{eq:final_quality_score}.}
	\label{tab:aucs_w}
	\resizebox{0.60\textwidth}{!}
	{
		\begin{tabu}{|[2pt]c|[2pt]c|c|c|c|c|c|c|[2pt]}

		\tabucline[2pt]{-}

		\multirow{2}{*}{\textbf{Dataset}} & \multicolumn{7}{c|[2pt]}{\textbf{Precision Curve AUCs}}\\

		\tabucline[1pt]{2-8}

		\textbf{} & \textbf{w=0.35} & \textbf{w=0.4} & \textbf{w=0.45} & \textbf{w=0.5} & \textbf{w=0.55} & \textbf{w=0.6} & \textbf{w=0.65}\\

		\tabucline[2pt]{-}

		\textbf{COB+OA} & 30.32\% & 31.15\% & 28.17\% & 29.67\% & 29.59\% & 28.37\% & 29.32\%\\ \hline
		
		\textbf{MCG+OA} & 26.66\% & 25.98\% & 25.44\% & 26.93\% & 26.40\% & 25.31\% & 26.08\%\\ \hline
		
		\textbf{RPN+OA} & 36.19\% & 36.30\% & 36.67\% & 35.97\% & 36.26\% & 35.87\% & 34.99\%\\ \hline

		\tabucline[2pt]{-}

		\end{tabu}
	}
\end{table*}

\begin{table*}
	\centering
	\huge
	\caption{Comparison of AUCs of precision and recall curves between models with and without the integration of our proposal objectness assessment network.}
	\label{tab:aucs_score}
	\resizebox{1.00\textwidth}{!}
	{
		\begin{tabu}{|[3pt]c|[3pt]c|c|c|c|c|c|[3pt]c|c|c|c|c|c|[3pt]}

		\tabucline[3pt]{-}

		\multirow{2}{*}{\textbf{Dataset}} & \multicolumn{6}{c|[3pt]}{\textbf{Precision Curve AUCs}} & \multicolumn{6}{c|[3pt]}{\textbf{Recall Curve AUCs}}\\

		\tabucline[1pt]{2-13}

		\textbf{} & \textbf{COB} & \textbf{COB+OA} & \textbf{MCG} & \textbf{MCG+OA} & \textbf{RPN} & \textbf{RPN+OA} & \textbf{COB} & \textbf{COB+OA} & \textbf{MCG} & \textbf{MCG+OA} & \textbf{RPN} & \textbf{RPN+OA}\\

		\tabucline[3pt]{-}

		\textbf{VOC2012} & 8.24\% & 31.15\% & 19.36\% & 26.93\% & 17.79\% & 36.67\% & 36.52\% & 40.94\%         & 28.23\% & 31.15\% & 24.24\% & 36.52\% \\ \hline
		\textbf{VOC2007} & 7.57\% & 29.12\% & 18.01\% & 25.05\% & 16.84\% & 34.49\% & 31.52\% & 35.73\%         & 24.81\% & 27.30\% & 21.54\% & 28.28\% \\ \hline
		\textbf{SOS} & 9.03\% & 30.34\% & 21.62\% & 28.02\% & 11.71\% & 27.81\% & 61.95\% & 69.87\%         & 53.97\% & 58.10\% & 27.06\% & 43.61\% \\ \hline

		\tabucline[3pt]{-}

		\end{tabu}
	}
\end{table*}

\begin{table*}
	\centering
	\huge
	\caption{Comparison of AUCs of precision and recall curves between models trained with different data preparations. \textbf{bN}: number of proposal windows (per image) generated during training.}
	\label{tab:bboxN}
	\resizebox{1.00\textwidth}{!}
	{
		\begin{tabu}{|[3pt]c|[3pt]c|c|c|c|c|c|[3pt]c|c|c|c|c|c|[3pt]}

		\tabucline[3pt]{-}

		\multirow{2}{*}{\textbf{Dataset}} & \multicolumn{6}{c|[3pt]}{\textbf{Precision Curve AUCs}} & \multicolumn{6}{c|[3pt]}{\textbf{Recall Curve AUCs}}\\

		\tabucline[1pt]{2-13}

		\textbf{} & \textbf{bN=2000} & \textbf{bN=1500} & \textbf{bN=1000} & \textbf{bN=500} & \textbf{bN=100} & \textbf{bN=50} & \textbf{bN=2000} & \textbf{bN=1500} & \textbf{bN=1000} & \textbf{bN=500} & \textbf{bN=100} & \textbf{bN=50}\\

		\tabucline[3pt]{-}

		\textbf{VOC2012} & 31.28\% & 31.18\% & 31.15\% & 30.83\% & 28.53\% & 27.99\% & 40.44\% & 40.97\% & 40.94\% & 40.19\% & 40.02\% & 40.11\%\\ \hline
		\textbf{VOC2007} & 29.10\% & 29.99\% & 29.12\% & 28.73\% & 27.58\% & 26.71\% & 36.15\% & 35.90\% & 35.73\% & 35.54\% & 34.68\% & 34.93\%\\ \hline
		\textbf{SOS} & 30.19\% & 29.92\% & 30.34\% & 30.08\% & 28.51\% & 27.59\% & 70.16\% & 69.77\% & 69.87\% & 69.76\% & 69.22\% & 68.92\%\\ \hline

		\tabucline[3pt]{-}

		\end{tabu}
	}
\end{table*}

\begin{table*}
	\centering
	\huge
	\caption{The effect of non-maximal suppression (NMS) on original object proposal generators. \textbf{I}: threshold of intersection-over-union (IOU) used in NMS. \textbf{-}: no NMS is used.}
	\label{tab:nms_rpn}
	\resizebox{1.00\textwidth}{!}
	{
		\begin{tabu}{|[3pt]c|[3pt]c|c|c|c|c|c|c|c|[3pt]c|c|c|c|c|c|c|c|[3pt]}

		\tabucline[3pt]{-}

		\multirow{2}{*}{\textbf{Dataset}} & \multicolumn{8}{c|[3pt]}{\textbf{Precision Curve AUCs (COB)}} & \multicolumn{8}{c|[3pt]}{\textbf{Precision Curve AUCs (RPN)}}\\

		\tabucline[1pt]{2-17}

		\textbf{} & \textbf{--} & \textbf{I=0.9} & \textbf{I=0.8} & \textbf{I=0.7} & \textbf{I=0.6} & \textbf{I=0.5} & \textbf{I=0.4} & \textbf{I=0.3} & \textbf{--} & \textbf{I=0.9} & \textbf{I=0.8} & \textbf{I=0.7} & \textbf{I=0.6} & \textbf{I=0.5} & \textbf{I=0.4} & \textbf{I=0.3}\\

		\tabucline[3pt]{-}

		\textbf{VOC2012} & 8.24\% & 7.72\% & 6.93\% & 6.30\% & 5.89\% & 5.60\% & 5.32\% & 5.03\% & 19.36\% & 16.06\% & 11.97\% & 8.61\% & 6.85\% & 6.19\% & 6.55\% & 7.91\%\\ \hline
		\textbf{VOC2007} & 7.57\% & 7.15\% & 6.49\% & 5.97\% & 5.62\% & 5.33\% & 5.09\% & 4.85\% & 18.01\% & 15.17\% & 11.17\% & 8.13\% & 6.44\% & 5.92\% & 6.35\% & 7.63\%\\ \hline
		\textbf{SOS} & 9.03\% & 8.44\% & 7.52\% & 6.89\% & 6.47\% & 6.16\% & 5.84\% & 5.57\% & 21.62\% & 10.77\% & 8.09\% & 5.83\% & 4.50\% & 3.91\% & 3.95\% & 4.51\%\\ \hline

		\tabucline[3pt]{-}

		\end{tabu}
	}
\end{table*}

\begin{table*}
	\centering
	\huge
	\caption{Comparison of AUCs of precision and recall curves of applying non-maximal suppression on object proposals re-ranked by our OA method. \textbf{I}: threshold of intersection-over-union (IOU) used in NMS. \textbf{-}: no NMS used.}
	\label{tab:nms_OA}
	\resizebox{1.00\textwidth}{!}
	{
		\begin{tabu}{|[3pt]c|[3pt]c|c|c|c|c|c|c|c|[3pt]c|c|c|c|c|c|c|c|[3pt]}

		\tabucline[3pt]{-}

		\multirow{2}{*}{\textbf{Dataset}} & \multicolumn{8}{c|[3pt]}{\textbf{Precision Curve AUCs}} & \multicolumn{8}{c|[3pt]}{\textbf{Recall Curve AUCs}}\\

		\tabucline[1pt]{2-17}

		\textbf{} & \textbf{--} & \textbf{I=0.9} & \textbf{I=0.8} & \textbf{I=0.7} & \textbf{I=0.6} & \textbf{I=0.5} & \textbf{I=0.4} & \textbf{I=0.3} & \textbf{--} & \textbf{I=0.9} & \textbf{I=0.8} & \textbf{I=0.7} & \textbf{I=0.6} & \textbf{I=0.5} & \textbf{I=0.4} & \textbf{I=0.3}\\

		\tabucline[3pt]{-}

		\textbf{VOC2012} & 31.15\% & 24.73\% & 16.93\% & 11.76\% & 9.19\% & 7.96\% & 7.26\% & 6.78\% & 40.94\% & 45.69\% & 48.90\% & 49.45\% & 46.40\% & 40.49\% & 35.48\% & 32.06\%\\ \hline
		\textbf{VOC2007} & 29.12\% & 23.54\% & 16.23\% & 11.41\% & 8.95\% & 7.85\% & 7.23\% & 6.80\% & 35.73\% & 40.66\% & 43.87\% & 44.47\% & 41.58\% & 36.36\% & 32.08\% & 29.34\%\\ \hline
		\textbf{SOS} & 30.34\% & 23.86\% & 16.40\% & 11.20\% & 8.80\% & 7.63\% & 7.00\% & 6.45\% & 69.87\% & 73.72\% & 77.73\% & 77.87\% & 73.23\% & 63.97\% & 56.52\% & 50.51\%\\ \hline

		\tabucline[3pt]{-}

		\end{tabu}
	}
\end{table*}

\subsection{Non-maximal Suppression}\label{subsec:nms}

Non-maximal suppression (NMS) is a useful technique for removing similar proposal boxes. In this work, several NMS-related experiments are conducted. First, we check how NMS will affect the performance of original proposal methods. Second, we apply NMS on the re-ranked proposals by our assessment network and check the performance again.

\subsubsection{NMS on Object Proposals Generated Using Existing Methods}

Table~\ref{tab:nms_rpn} shows the performance when NMS with different intersection-over-union (IOU) thresholds are applied on object proposals. Here we use COB~\cite{cob-eccv2016-maninis} and RPN~\cite{faster_rcnn-nips15-ren} for proposal generation. It can be seen clearly that both precision of COB and RPN drop quickly as the threshold decreases (Table~\ref{tab:nms_rpn}). This is due to that some inaccurate proposal boxes are falsely rated with high scores, misleading NMS to remove many high-objectness proposals near them. This is not desirable by our method, since it does not create new proposals. If the proposals are generally of low objectness, it is less-likely that our method can do any better. In this work, we recommend using NMS with larger thresholds (such as $>=0.7$) on proposals before performing assessment on object proposals.

\subsubsection{NMS on Object Proposals Re-ranked by Proposal Objectness Assessment}

Table~\ref{tab:nms_OA} shows the performance of our method, with NMS being applied on re-ranked proposals. Compared to COB and RPN, our model still clearly outperforms them under the same IOU thresholds. When the threshold decreases, the precision drops quickly. The selection of the IOU threshold is empirical and subject to specific applications. In our pipeline of object mining, we use the threshold of 0.5 in building the object database.

\subsection{User Studies}

Although quantitative results in the previous section have demonstrated the effectiveness of our method, it still needs to be confirmed that such improvement does have positive impact on obtained object database and visual searching results.

From the same image collection downloaded from Flickr, we build three object databases using our method and other state-of-the-art object proposal methods. As we are more concerned about precision, we choose two existing object proposal methods with good precision performance, MCG~\cite{mcg-cvpr2014-arbelaez} and RPN~\cite{faster_rcnn-nips15-ren}. For MCG and RPN, top 50 proposal windows are generated for each image and are screened by non-maximum suppression, removing duplicated objects. For our method, the object mining pipeline described in Section~\ref{sec:overview} is used.

During this user study, each user is given a series of test cases. Within each test case, a query image is used to retrieve top-ranked 50 images from each of the three databases. Three groups of retrieved images are shown to each user in a random order, and the user is asked to pick all ``good" objects. For the selection of ``good" objects, the user is told to select those images having objects in good condition (satisfying {\em completeness} and {\em fullness}). We choose 30 query images as 30 test cases, which are evenly distributed to 30 participants. Each participant receives 3 test cases consisting of 9 groups of images in total. Among all participants, 15 people have certain level of knowledge of image editing, while the rest 15 people have not.

Given a participant and an object database, we quantify the score of the database as the percentage of objects picked by user, 

\begin{equation}
N_{picked}/N_{total}
\end{equation}

where $N_{picked}$ and $N_{total}$ are the number of objects picked by the participant and the total objects within the database, respectively. The quantity indicates the ``purity'' of good objects within a given database. For all three databases, we calculate their averaged scores over results obtained from 30 participants. According to records of the user study. The database built using our method has significantly better quality. Specifically, 57.58\% of objects from our database are picked in comparison to 34.50\% for MCG and 19.15\% for RPN.


\begin{figure*}
	\centering
	
	\includegraphics[width=1.0\linewidth]{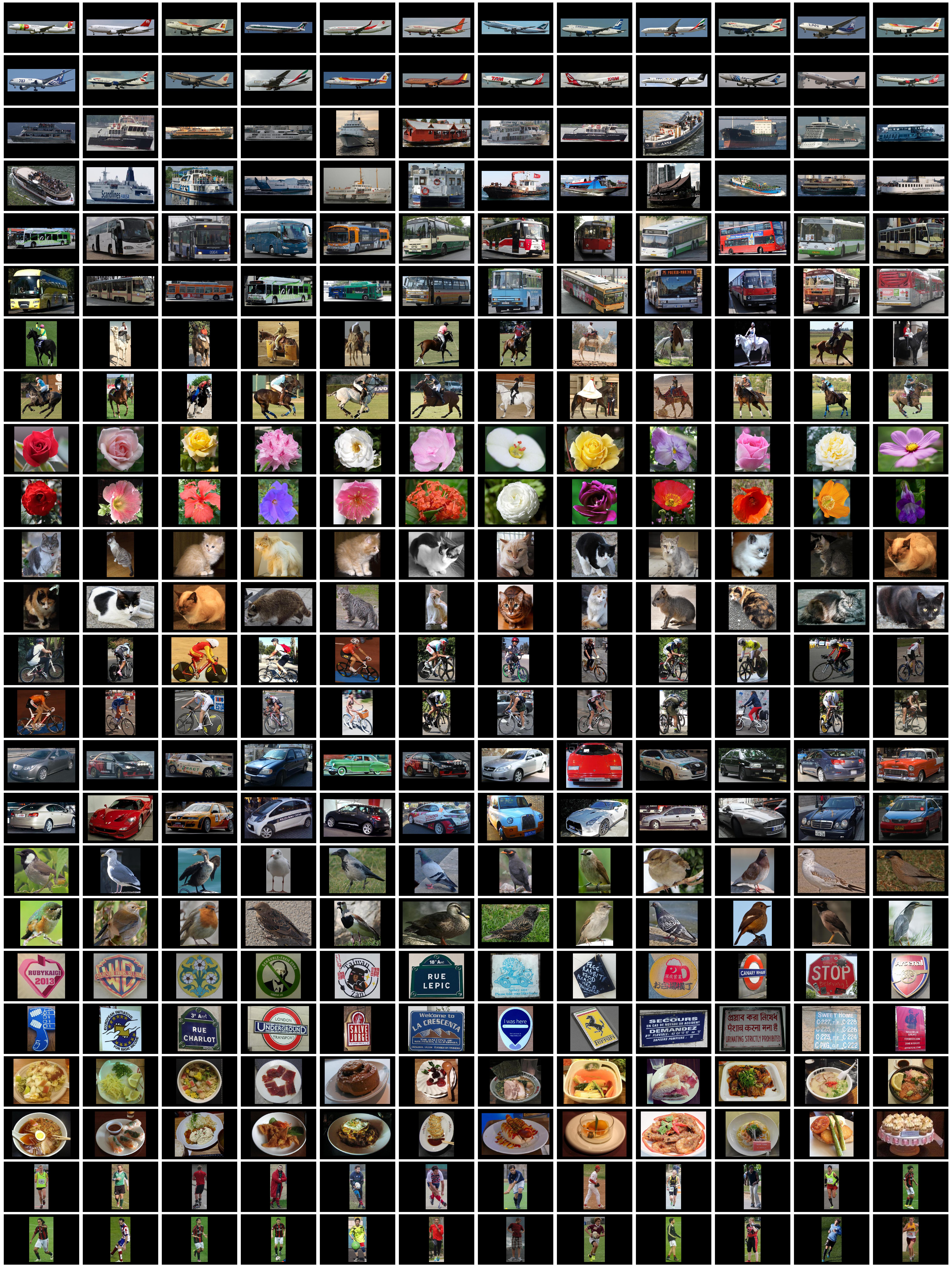}
	
	\caption{A sample collection of visual objects discovered using our pipeline. Every row corresponds to a group of visually similar objects. Note that our method does not assume object categories. The grouping of similar objects is through K-means over deep features~\cite{vgg16-corr2014-simonyan} of the chosen object boxes.}
	\label{fig:object_grid}
	
\end{figure*}

\begin{figure*}
	\centering
	
	\includegraphics[width=1.0\linewidth]{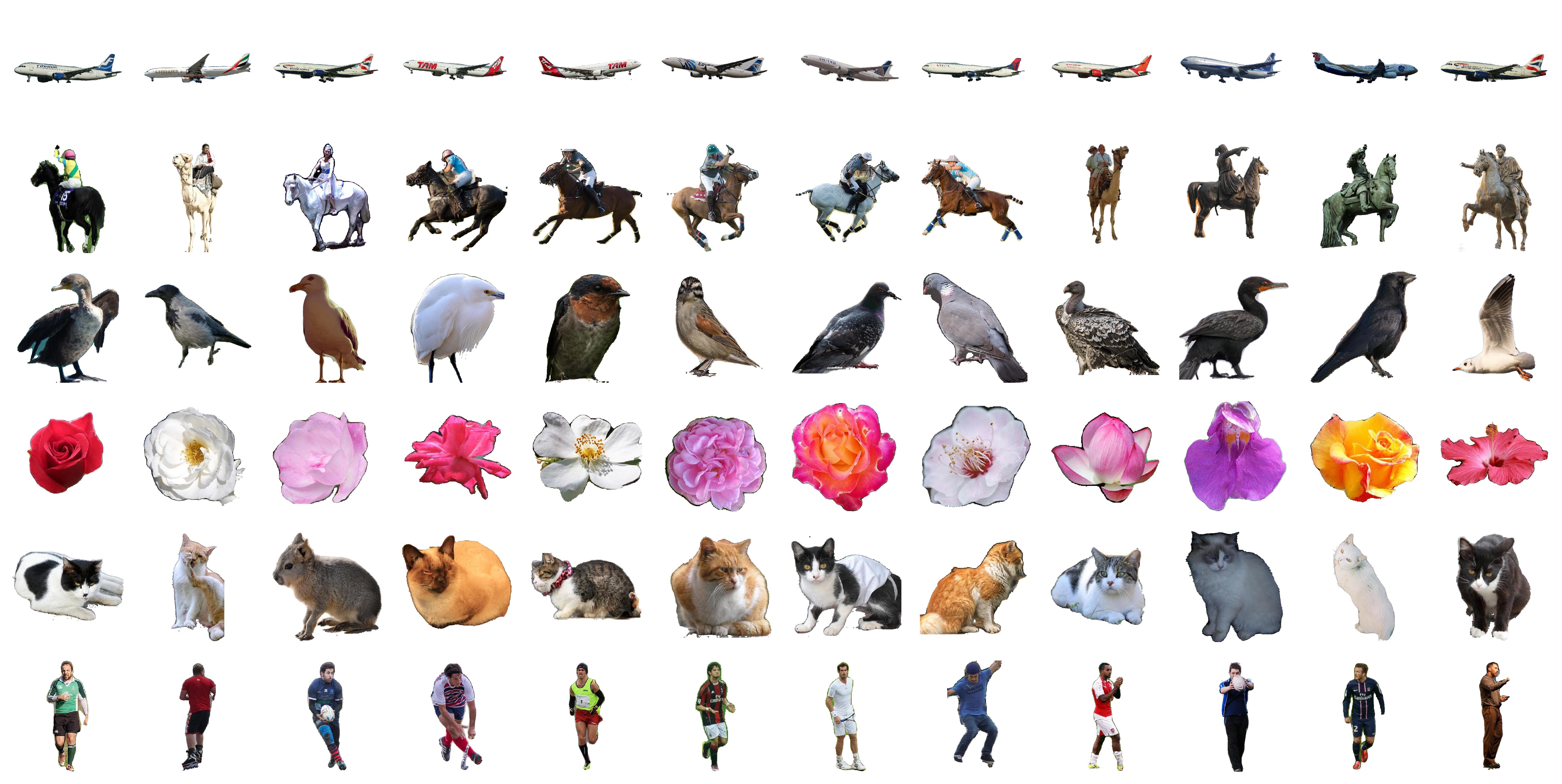}
	
	\caption{Results of applying automatic segmentation on some of the extracted objects from Flickr images. Method described in \cite{auto_obj_seg-iet_ipr_2018-wu} is used.}
	\label{fig:object_seg}
	
\end{figure*}

\section{Application}\label{sec:app}

\subsection{Accessing Object Database}

Figure~\ref{fig:object_grid} shows some of the objects obtained from Flickr using our method. Since our method performs class-agnostic visual object discovery and locating from internet images, objects in our object database are not associated with any specific category labels. Therefore, locating a desired object relies on visual search. To make our object database easy to access, we provide two types of query interfaces, one is content based image retrieval, and the other is sketch-based image retrieval.

\begin{figure}
	\centering
	
	\begin{subfigure}[]{0.24\linewidth}
		\includegraphics[width=\linewidth]{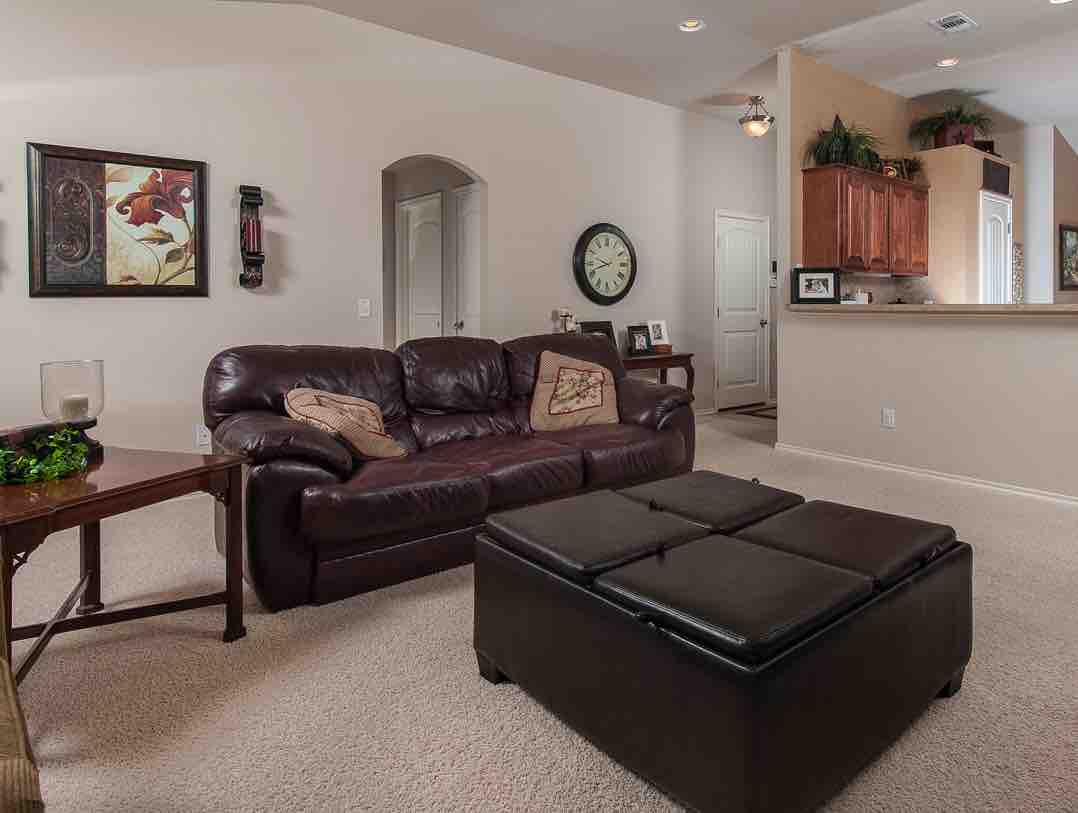}
		\caption{}
		\label{fig:09_01_01}
	\end{subfigure}
	\begin{subfigure}[]{0.24\linewidth}
		\includegraphics[width=\linewidth]{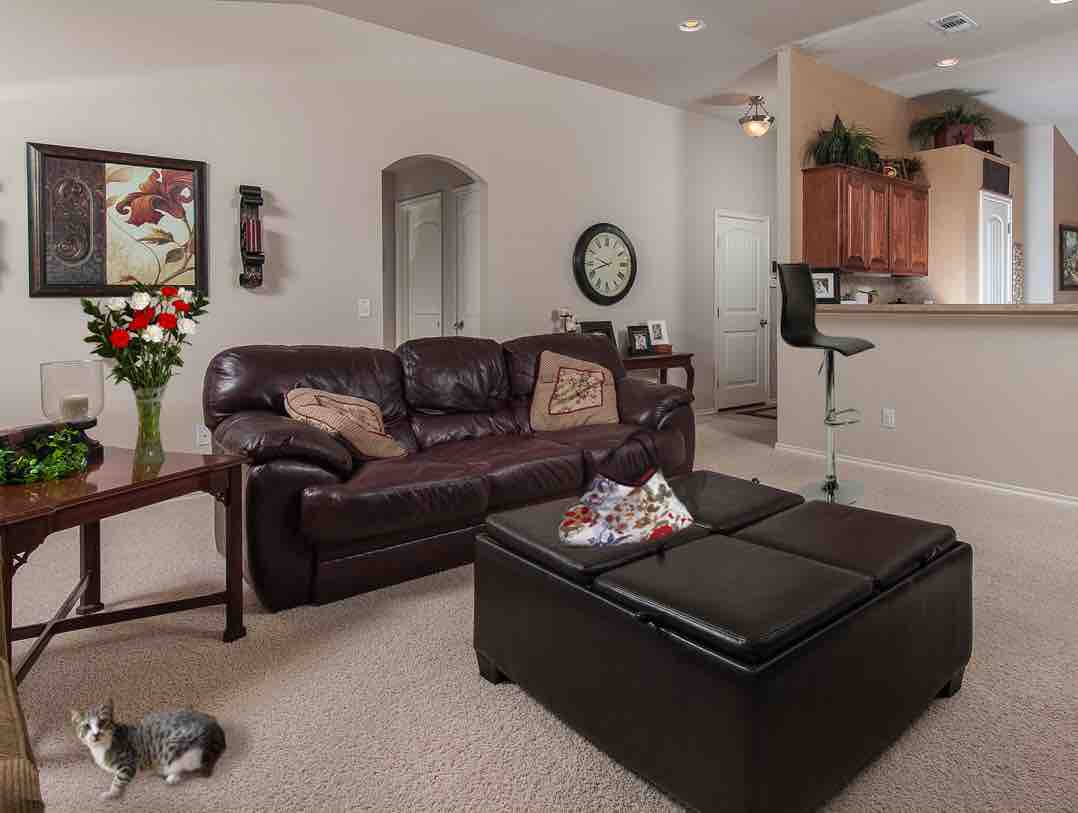}
		\caption{}
		\label{fig:09_01_02}
	\end{subfigure}
	\begin{subfigure}[]{0.24\linewidth}
		\includegraphics[width=\linewidth]{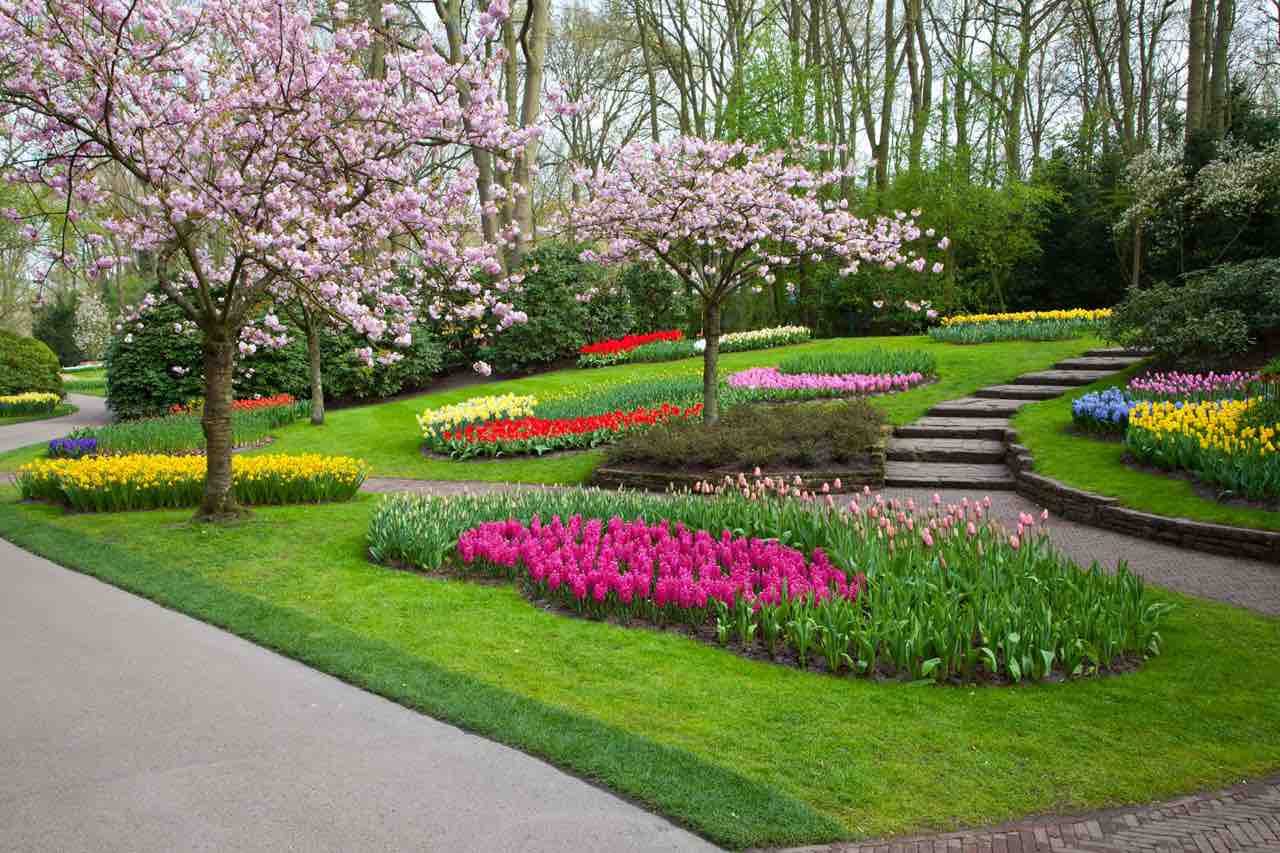}
		\caption{}
		\label{fig:09_02_01}
	\end{subfigure}
	\begin{subfigure}[]{0.24\linewidth}
		\includegraphics[width=\linewidth]{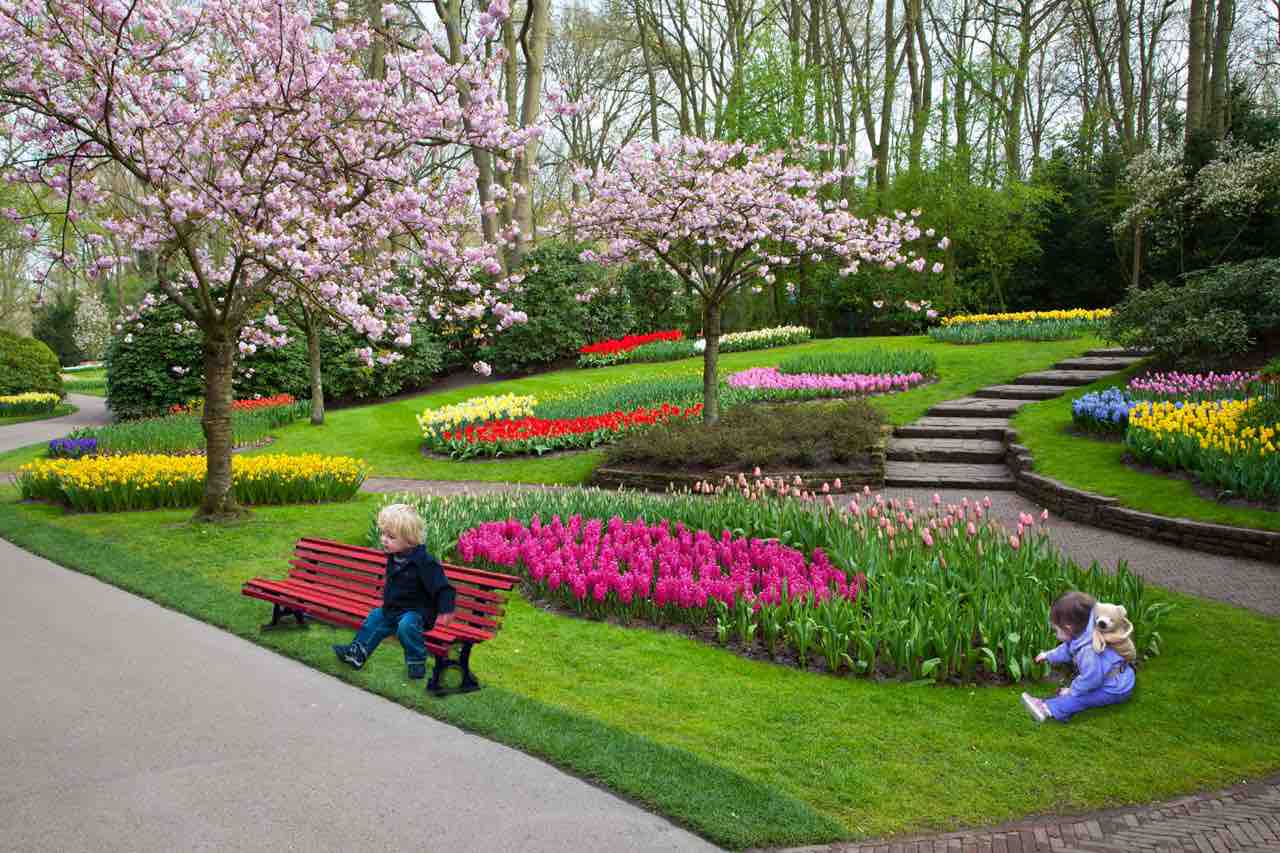}
		\caption{}
		\label{fig:09_02_02}
	\end{subfigure}

	\begin{subfigure}[]{0.24\linewidth}
		\includegraphics[width=\linewidth]{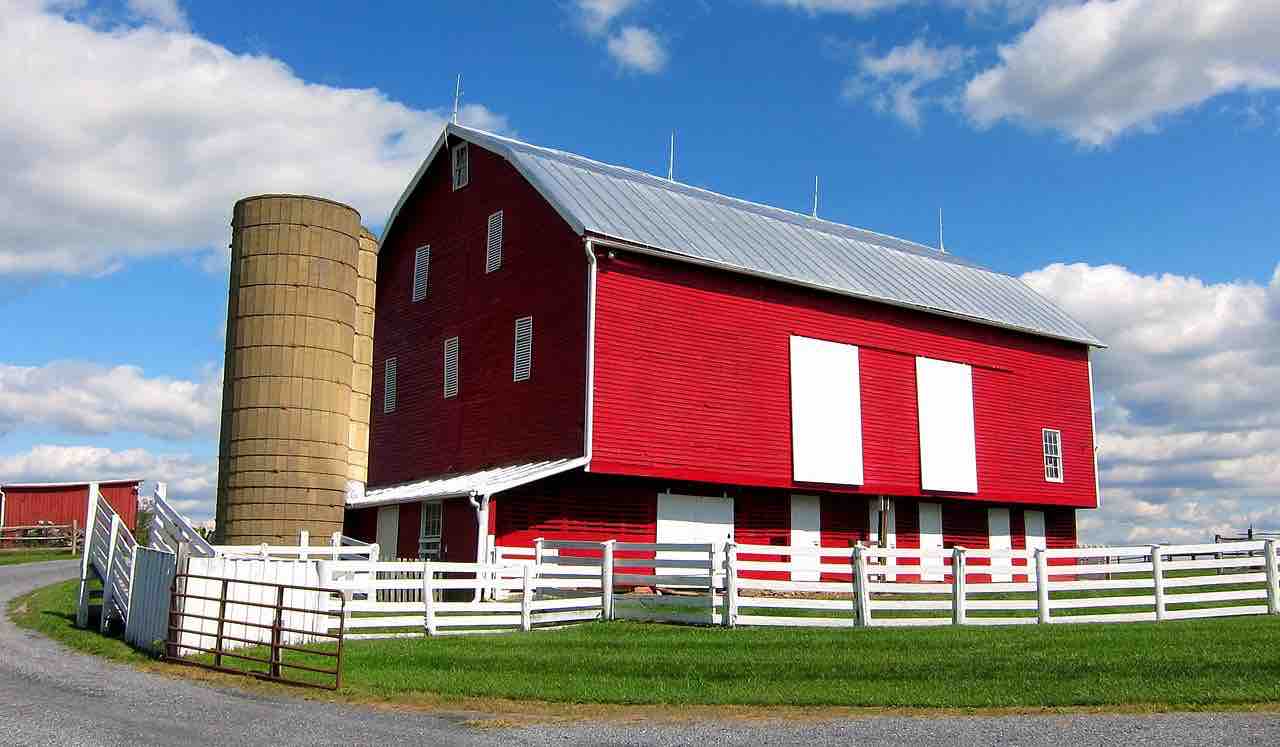}
		\caption{}
		\label{fig:09_03_01}
	\end{subfigure}
	\begin{subfigure}[]{0.24\linewidth}
		\includegraphics[width=\linewidth]{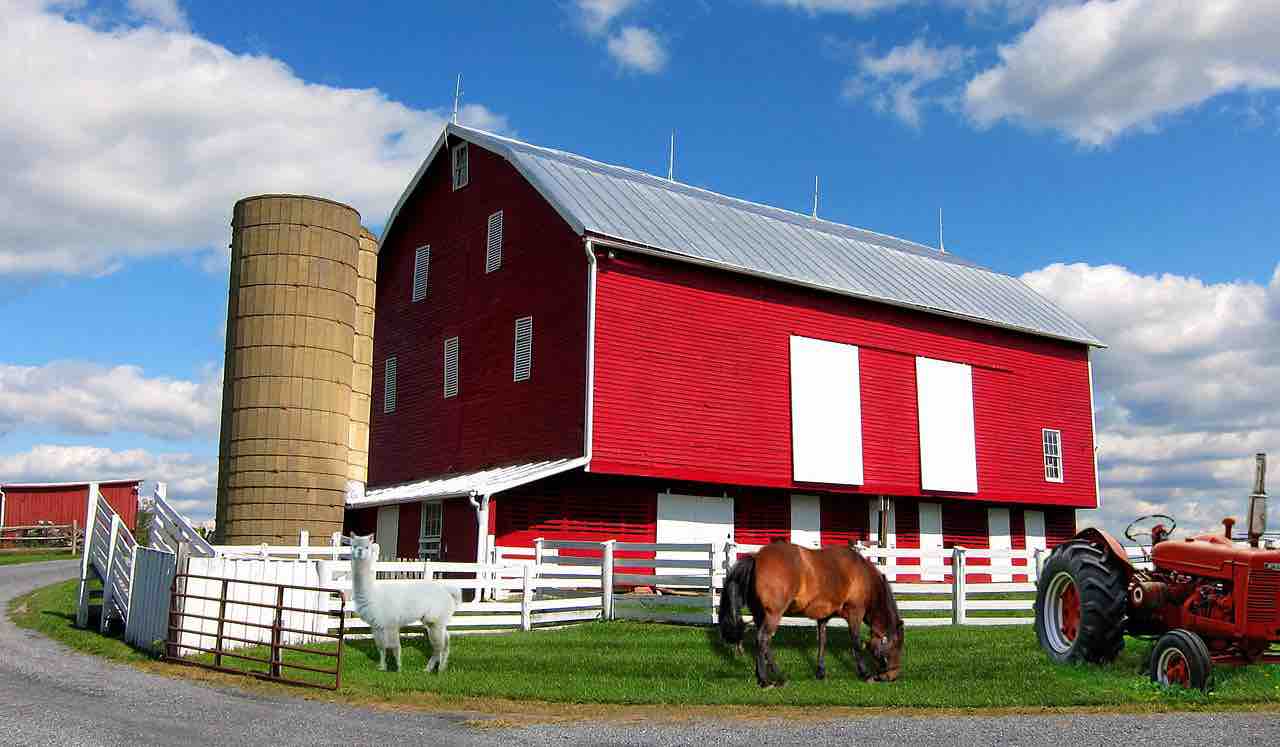}
		\caption{}
		\label{fig:09_03_02}
	\end{subfigure}
	\begin{subfigure}[]{0.24\linewidth}
		\includegraphics[width=\linewidth]{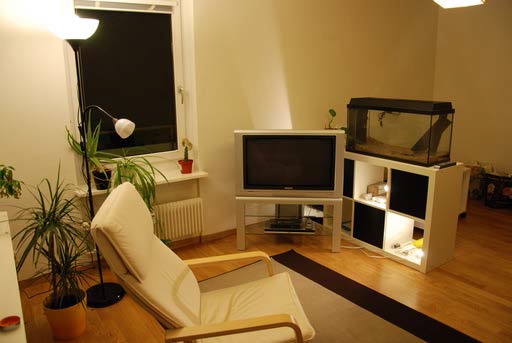}
		\caption{}
		\label{fig:09_04_01}
	\end{subfigure}
	\begin{subfigure}[]{0.24\linewidth}
		\includegraphics[width=\linewidth]{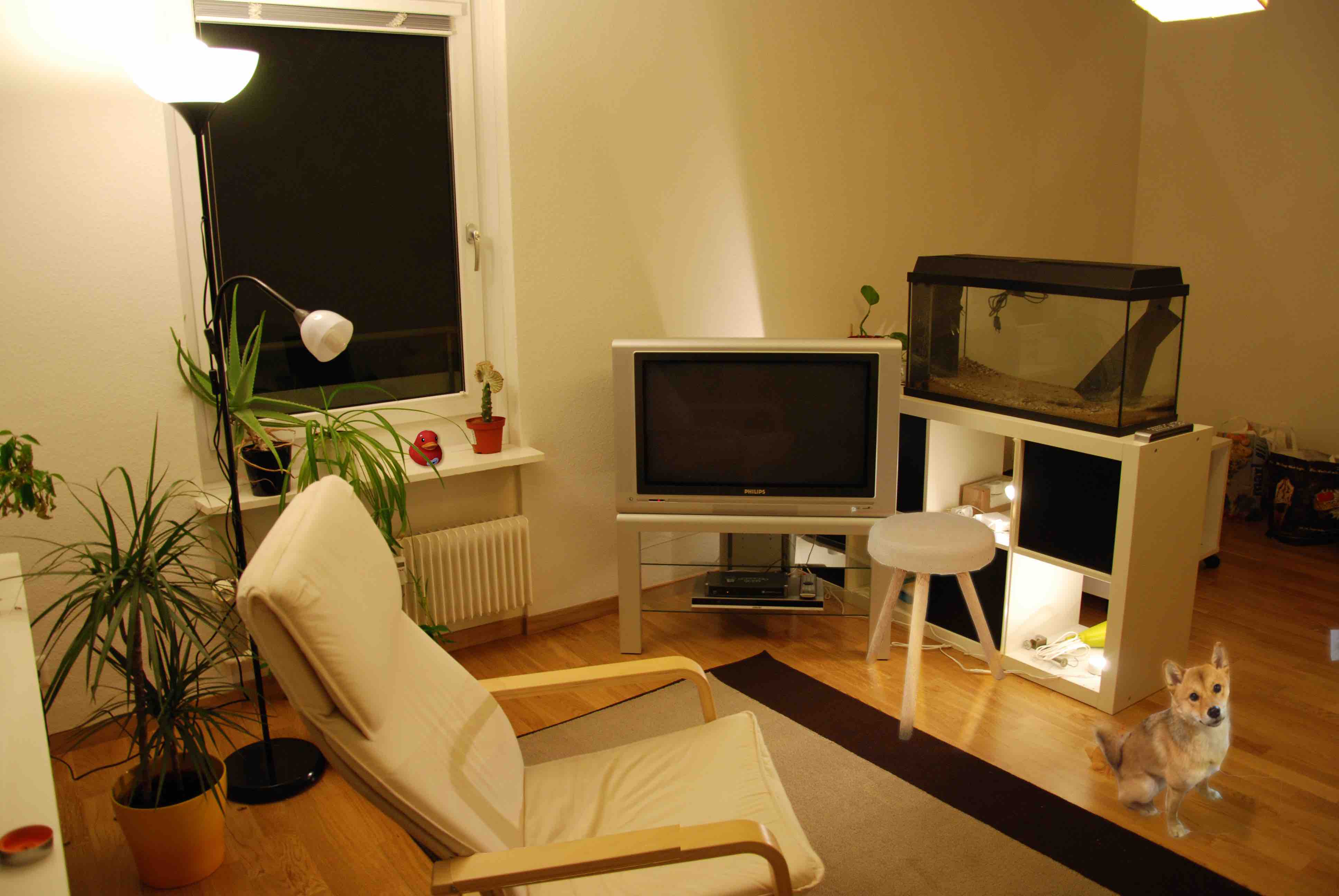}
		\caption{}
		\label{fig:09_04_02}
	\end{subfigure}
	
	\caption{Image compositing examples using objects from our database. Adobe Photoshop is used here to make blended results more realistic.}

	\label{fig:09_01_image_editing}

\end{figure}

For the interface of content based image retrieval, the user first supplies an image of an object, which is used to retrieve a list of images containing similar objects. Image retrieval is implemented using deep feature extraction with a pre-trained VGG16 network~\cite{vgg16-corr2014-simonyan} and nearest neighbor searching based on the cosine distance metric~\cite{cbir-icm2014-wan, cbir-cvpr2014-wang}. If a background image is also provided, we can further apply a refinement process to re-rank retrieved visual objects with respect to the context of background image. In this step, the background image is used as the query image to re-rank already retrieved image list.

For the interface of sketch-based object retrieval, our system provides a panel for the user to draw a sketch. Our system searches the database to retrieve a list of objects which can be associated with the sketch. Sketch-based image retrieval is based on the deep CNN model provided in \cite{sangkloy2016sketchy}. Similarly, given a background image, our system can also re-rank retrieved objects using the same method regarding background similarities, as provided in the first type of user interface.

\subsection{Image Compositing}

Image compositing can be performed conveniently using our object database. Here we look at an application of blending objects into a given background image.

Given a background image and some foreground object images found within our database, we firstly apply automatic or semi-automatic image segmentation (such as \cite{grabcut-siggraph2004-rother}) on foreground object images, and alpha matting (such as~\cite{knn_matting-tpami-chen}) to cut out foreground objects. We further apply improved Poisson blending and alpha blending as proposed in \cite{sketch2photo-sigraphasia2009-chen} to generate a high-quality composition. Some image compositing examples are shown in Figure \ref{fig:09_01_image_editing}.

\subsection{Automatic Segmentation}

The selection of high objectness proposals can be potentially useful in fully-automatic segmentation. Figure~\ref{fig:object_seg} shows some of the results on object images obtained from Flickr. The method from~\cite{auto_obj_seg-iet_ipr_2018-wu} is used. Please note that although not all object images obtained from Flickr can be processed well by this method, there are still a large number of acceptable results.

\section{Conclusions and Discussion}

In this paper, we present an effective pipeline for discovering and locating visual objects from internet images. This pipeline is based on dense object proposal generation and objectness assessment. A deep neural network is designed for the inference of proposal objectness score, based on the criteria of {\em completeness} and {\em fullness}. Objectness scores returned from this network are used to re-rank pre-generated object proposals to choose optimal object proposal windows. Our experiments confirm the effectiveness of the proposed method, showing its relatively higher precision and recall when compared to existing state-of-the-art methods.

From the perspective of building an object database, our pipeline and the assessment network can be useful in many applications, such as building a photo gallery of visual objects for a specific image collection, performing image editing with diversified choices of visual objects, enriching existing public visual object databases, etc. More interestingly, the proposal objectness assessment can potentially be useful in improving the performance of an object detector, as proposals that are less-likely to contain any objects can be safely removed using this method. This could be an interesting future work.

\textbf{Limitation} Our method is dependent on the performance of existing object proposal generators. If an object is not detected by the proposal generator, our method is not able to find it as well. In the future, we intend to combine object proposal generation and proposal objectness assessment in a unified framework, and improve the performance of proposal generation.


\end{document}